\pdfoutput=1
\documentclass[11pt]{article}

\usepackage[]{EMNLP2023}

\usepackage{times}
\usepackage{latexsym}

\usepackage[T1]{fontenc}
\usepackage[utf8]{inputenc}

\usepackage{microtype}

\usepackage{inconsolata}

\usepackage{csquotes}
\usepackage{booktabs}
\usepackage{multirow}
\usepackage{subfigure}
\usepackage{graphicx}
\usepackage{xcolor}
\usepackage{color}
\usepackage{amsmath}
\usepackage{amsthm}
\usepackage{amssymb}
\usepackage[capitalise,noabbrev]{cleveref}
\usepackage{adjustbox}

\makeatletter
\renewcommand\paragraph{%
  \@startsection{paragraph}
    {4}
    {\z@}
    {3.25ex \@plus1ex \@minus.2ex}
    {-1em}
    {\normalfont\normalsize\bfseries\maybe@addperiod}%
}
\newcommand{\maybe@addperiod}[1]{%
  #1\@addpunct{.}%
}
\makeatother

\newcommand{\sgquote}[1]{\textquoteleft #1\textquoteright}

\newcommand{\model}{}
\def\model/{KnowRec}

\def\smallcol{\hspace{6pt}}
\def\tinycol{\hspace{4pt}}
\title{Knowledge-grounded Natural Language Recommendation Explanation}

\author{\textbf{Anthony Colas}\textsuperscript{*1}, \textbf{Jun Araki}\textsuperscript{2}, \textbf{Zhengyu Zhou}\textsuperscript{2}, \\ \textbf{Bingqing Wang}\textsuperscript{2}, \textbf{Zhe Feng}\textsuperscript{2} \\
  \textsuperscript{1}University of Florida \\
  \textsuperscript{2}Bosch Research North America \\
  \texttt{acolas1@ufl.edu} \\
   \texttt{\{jun.araki, zhengyu.zhou2, bingqing.wang, zhe.feng2\}@us.bosch.com}}

\begin{document}
\maketitle
\newcommand\blfootnote[1]{%
\begingroup
\renewcommand\thefootnote{}\footnote{#1}%
\addtocounter{footnote}{-1}%
\endgroup
}
\blfootnote{*Work performed at Bosch Research.}

\begin{abstract}
Explanations accompanied by a recommendation can assist users in understanding the decision made by recommendation systems, which in turn increases a user's confidence and trust in the system. Recently, research has focused on generating natural language explanations in a human-readable format. Thus far, the proposed approaches leverage item reviews written by users, which are often subjective, sparse in language, and unable to account for new items that have not been purchased or reviewed before. Instead, we aim to generate fact-grounded recommendation explanations that are objectively described with item features while implicitly considering a user's preferences, based on the user's purchase history. To achieve this, we propose a knowledge graph (KG) approach to natural language explainable recommendation. Our approach draws on user-item features through a novel collaborative filtering-based KG representation to produce fact-grounded, personalized explanations, while jointly learning user-item representations for recommendation scoring. Experimental results show that our approach consistently outperforms previous state-of-the-art models on natural language explainable recommendation.
\end{abstract}
\section{Introduction}\label{sec:intro}
Current approaches to natural language (NL) explainable recommendation focus on generating user reviews
%, because of the data currently available to train such generative models
~\cite{chen2018neural,wang2018explainable,li2020generate,li-etal-2021-personalized, yang2021explanation}. Instead of providing a justification for the item recommendation, the models learn to output language that is commonly found in personal reviews. This reliance on reviews poses three problems: 1) The explanations are not objective, because users typically review items based on their sentiment~\cite{wu2018improving}, 2) Reviews are often sparse, because they describe a user's own experience~\cite{asghar2016yelp}, 3) Systems that rely on reviews cannot account for new items which have never been purchased before, nor can they provide justifications for item catalogs which may not have reviews
available. Given this, it may be difficult for a user to reason as to why an item was recommended, hindering the user's experience~\cite{tintarev2015explaining}. 
% Take for example an output from PETER~\cite{li-etal-2021-personalized}, an explainable recommendation system which uses aspect features found in reviews as input: \textit{\enquote{The lobby was very nice and the rooms were very comfortable.}} While the explanation is fluent and describes hotel-related features, such explanation is subjective, not specific to a given hotel, and relies on data from pre-existing reviewed items. \ac{Do we need this sentence if we the Paths of Glory example?}
The user may then lose trust in such systems which do not provide objective and accurate explanations.
%pertaining to the recommended items.

%talk about kg recommendation and how we can leverage this
%problem we are solving
%While previous work is on review-based explanation, 
%TODO: AC do we need this?
% To address the aforementioned problems, we introduce a new task to generate fact-grounded NL explanations that are objectively described with item features, along with a recommendation.
%We then propose a knowledge graph (KG) based approach that can successfully produce fact-grounded and personalized explanations with item features, being scalable to unpurchased products.
We propose \textbf{\model/}, a KG-grounded approach to natural language explainable recommendation which not only personalizes recommendations/explanations with user information, but also draws on facts about a particular item via its corresponding KG to generate objective, specific, and data-driven explanations for the recommended item. For example, given the movie \enquote{Paths of Glory}, previous work aims to generate explanations such as \enquote{it's not the best military movie} and \enquote{good performances all around}, which are subjective, not specific to a given movie, and relies on data from pre-existing reviews. Instead, by leveraging an item KG such as \textit{<director, Stanley Kubrick>, <conflict, World War 1>, <country, France>}, a more objective and precise explanation can be produced such as: \enquote{A World War I French colonel defends three soldiers. Directed by Stanley Kubrick.}  The item features of `World War I', `colonel', and 'defends three soldiers' in the explanation objectively describe the movie, while they can implicitly reflect the user's preferences for war movies, based on his/her purchase history.

\model/ is also more advantageous than prior work in terms of scalability to unpurchased items.
Previously, KG-based recommendation systems have effectively addressed the cold-start problem by linking users and items through shared attributes~\cite{wang2019kgat, wang2020ckan, wang2021learning}. Similarly, there exists a kind of cold-start problem for new items in recommendation explanation that rely on reviews. \model/ demonstrates KGs can help solve this problem through existing item-level features by adapting KG-to-text~\cite{koncel-kedziorski-etal-2019-text, ke-etal-2021-jointgt, colas-etal-2022-gap} elements into explainable recommendation, producing item-level explanations to justify a purchase. 
%While previous work focuses on review-based explanations, we introduce a new approach to NL explainable recommendation, which incorporates knowledge graphs (KGs) to produce objective and information rich descriptions which can scale to unpurchased products. \ac{made this more clear. Before it said: "introduce a new task to solve aforementioned problems"}
%KGs are fact-grounded data representations which consider each item and its specific attributes as nodes with corresponding relationships between them~\cite{tintarev2015explaining}. 
%Previously, KG-based recommendation systems have effectively addressed the cold-start problem by linking users and items through shared attributes~\cite{wang2019kgat, wang2020ckan, wang2021learning}. Similarly, there exists a kind of cold-start problem for new items in recommendation explanation which rely on reviews. KGs can help solve this problem through existing item-level features by adapting KG-to-text~\cite{koncel-kedziorski-etal-2019-text, ke-etal-2021-jointgt, colas-etal-2022-gap} elements into explainable recommendation, producing item-level explanations to justify a purchase. 
%as to why an item was recommended to a user as seen in \cref{fig:task}, where a user and item KG are integrated to produce an item specific recommendation and explanation.
The KG-based approach is particularly important for recommendation scenarios in special domains where personal reviews are not available and the review-based approaches are impractical.

Our approach presents several algorithmic novelties. First, inspired by work on KG Recommendation~\cite{wang2020ckan} and KG-to-Text~\cite{colas-etal-2022-gap}, we devise a novel user-item KG lexical representation, viewing the input through collaborative filtering lens, where users are graphically represented via their previous purchases and connected to a given item KG. Our representation differs from previous work on explainable NL generation which relies on ID and sparse keyword features. Previous work extracts keywords from reviews to represent the user and item, linearizing all such features to encode and produce an NL explanation~\cite{li2020generate,li2022personalized}.  Next, KnowRec adapts a graph attention encoder for the user-item representation via a new masking scheme. Finally, the encoded KG representation is simultaneously decoded into a textual explanation, while we innovatively dissociate the joint learned user-item representation to compute a user-item similarity for recommendation scoring. 
%\zz{Much better now. But as discussed in meeting, this and previous paragraphs need to further improved (state that we aim to address a new task, etc)}
%\zz{This paragraph should be re-organized to improve the clarity. The definition of the graph, i.e., "where users are represented via their purchase history..." should be put in an earlier place, i.e., before or right after the presentation of the figure. Another problem is that the use of terms "user", "item", "features" should be consistent across the Figure title and paragraph text. Currently there are many conflicts, making it hard to understand for readers. Whether or not we should describe the KG in such details in Introduction part is something that needs to be carefully considered. While here a detailed description is provided, the contributions of our algorithm are not sufficiently highlighted.} \ac{I have rewritten this paragraph for clarity and to talk more about the algorithm. Thank you.}%

To evaluate our approach, we first devise a method of constructing $(KG, Text)$ pairs from product descriptions as described in \cref{sec:dataset}, where we extract entities and relations for the item KGs. We construct two such datasets from the publically available recommendation datasets to evaluate our proposed model for both the explanation and recommendation task and focus on natural language generation (NLG) metrics for the explanation task as in previous work. We adapt and compare previous baseline models for the recommendation explanation task as described in \cref{sec:experiments}, where we substantially outperform  previous models on explanation while achieving similar recommendation performance as models that rely on user and item ID-based features.
% \footnote{We will release our code and datasets upon acceptance.}

%Maybe Contributions:
%1. Point out shortcomings in current generative explainable recommendation
%1. Devise KG-explainable recommendation
%2. Propose Dataset
%3. Propose model
%5. Experimental results

%figure showing overview of task (KG Rex Sys)

\section{Related Work}\label{sec:related}
\begin{figure*}[t]
\centering
\includegraphics[width=\textwidth]{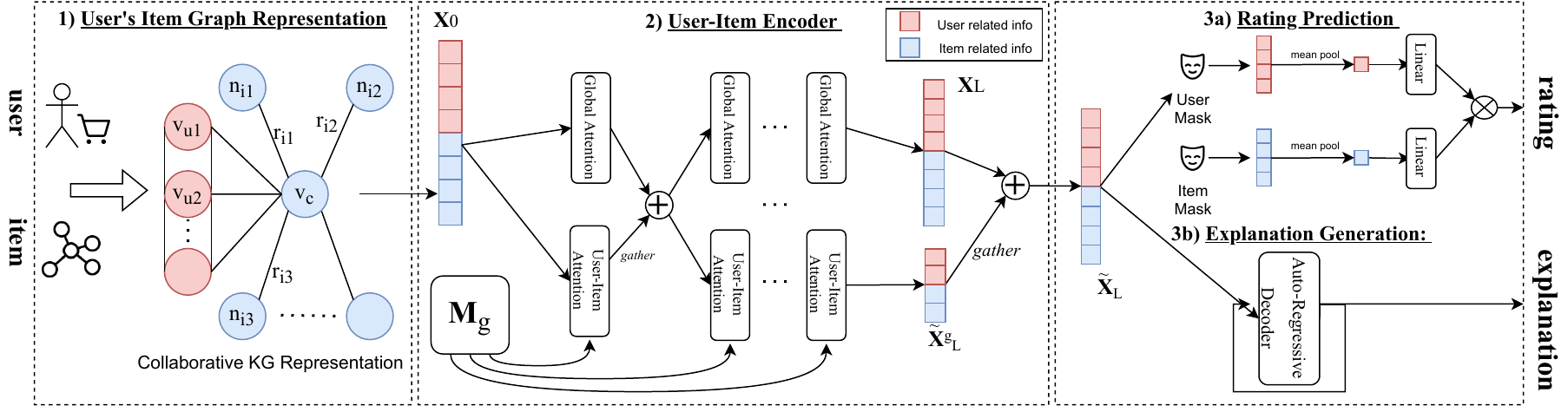}
\caption{Illustration of KnowRec. 1) The User's Item KG Representation Module. 2) The Global and User-Item Graph Attention Encoder. 3) The Output Module for rating prediction and explanation.
}
%\zz{it will be more clear to replace the small graph picture (i.e., the one output by the last User-Intent attention encoder) with a vector representation, which is similar as X-L but notated as X tilter-gL. }
%\jun{Alternatives to user-item KG: (1) user's Item Graph Representation and (2) items-per-user Graph Representation.} \ac{Have replaced it with (1)} \jun{Do we need $\mathbf{X}_L$ twice in the middle? If the two vectors $\tilde{\mathbf{X}}_L$ are the same, can we have just one to remove the duplicate and save space? Please use $\mathbf{W}^u$ and $\mathbf{W}^v$ (superscripts not in bold font), along with \cref{eq:rating}.}\ac{Thank you for correctig those typos, I have put the new image to reflect those changes.}
\label{fig:model}
\end{figure*}
\subsection{Explainable Recommendation}
Previous works on NL explainable recommendation focus on generating user-provided reviews, where the output is typically short, subjective, and repetitive~\cite{chen2018neural, hou2019explainable, wang2018reinforcement, yang2021explanation, li2017neural, li2020generate, li-etal-2021-personalized, hui2022personalized}.
%Extractive approaches~\cite{chen2018neural, hu2008collaborative, wang2018reinforcement}  focus on directly choosing text segments from existing reviews, while 
Extractive-based approaches have been proposed to score and select reviews as explanations~\cite{chen2018neural, li2019capsule}. Conversely, generative approaches~\cite{yang2021explanation, li2017neural, li2020generate, li-etal-2021-personalized, 10.1145/3366423.3380164, hui2022personalized} leverage user/item features to generate new reviews as explanations.
% In PETER, a transformer~\cite{vaswani2017attention} encoder is modified with a user and item embedding for the explainable recommendation task. 
Currently, the task is still limited by review data, thus these models cannot adequately handle new items.
% Additionally, these generative approaches rely on sparse user-item~\cite{li2020generate} and aspect-based features~\cite{li-etal-2021-personalized}. 
Unlike previous work, we introduce KGs to the explainable recommendation task to provide objective, information-dense, specific explanations. Our approach can then handle new items which have not been reviewed yet.

Inspired by recent advancements in explainable recommendation models like~\cite{li-etal-2021-personalized}, we enhance BART \cite{lewis-etal-2020-bart}, renowned for graph-to-text tasks, to incorporate user-item knowledge graphs. This adaptation enables us to generate recommendation scores along with natural language explanations.

\subsection{Knowledge Graph Recommendation}
Leveraging KGs for recommendation systems has gained increasing attention~\cite{wang2019kgat, wang2020ckan, wang2021learning, xie2021explainable, du2022hakg}. In neighborhood-based methods~\cite{hamilton2017inductive, welling2016semi, velivckovic2018graph}, propagation is performed iteratively over the neighborhood information in a KG to update the user-item representation. While recent work has produced explanations via KGs, these works focus on structural explanations such as knowledge graph paths~\cite{ma2019jointly, fu2020fairness, xian2019reinforcement} and rules~\cite{zhu-etal-2021-faithfully, chen2021neural,shi2020neural}, which are not as intuitive for users to understand. We focus on generating NL explanations, which has been shown to be a preferred type of explanation~\cite{zhang2020explainable}. For a fair comparison, we compare to prior work that produces NL explanations. Unlike these works, we aim to generate NL explanations instead of using paths along the KG as explanations. 

\subsection{Knowledge Graph-to-Text Generation}
In KG-to-Text, pre-trained language models such as GPT-2~\cite{radford2019language} and BART~\cite{lewis-etal-2020-bart} have seen success in generating fluent and accurate verbalizations of KGs~\cite{chen-etal-2020-kgpt, ke-etal-2021-jointgt, ribeiro-etal-2021-investigating, colas-etal-2022-gap}. %Similarly, we base our model on BART, to take full advantage of the pre-trained model's generation capabilities. 
%While our model is most similar to GAP, where a graph encoder is appended to BART, its generation settings and masking scheme are formulated for the explainable recommendation task. 
We devise an encoder for user-item KGs and a decoder for both the generation and recommendation tasks. Specifically, we formulate a novel masking scheme for user-item KGs to structurally encode user and item features, while generating a recommendation score from their latent representations. Thus, our task is two-fold, fusing elements from the Graph-to-Text generation and KG recommendation domains.

\section{Problem Formulation}\label{sec:problem}
Following prior work, we denote $\mathcal{U}$ as a set of users, $\mathcal{I}$ as a set of items, and 
%$\mathcal{U}$ as a set of users, $\mathcal{I}$ as a set of items, and 
the user-item interaction matrix as $\mathbf{Y} \in \mathbb{R}^{|\mathcal{U}| \times |\mathcal{I}|}$, where $y_{uv} = 1$ if user $u \in \mathcal{U}$ and item $v \in \mathcal{I}$ have interacted.
%\jun{Are $\mathbf{Y}$ and $y_{uv}$ used in any other places?  If not, we do not necessarily need to introduce those notations.}\ac{I commented out $\mathbf{Y}$, but we use $y_{uv}$ in equation (1)}\jun{Okay, $\mathbf{Y}$ also needs to be introduced here because it is referred to by \cref{sec:encoder}.}\ac{Typo, Y is mentioned in the previous senence}
%\jun{Please check if this is correct.}\ac{Thank you it's correct}.
Here, we represent user $u$ as the user's purchase history $u = \{ v_{ui} \}$, where $v_{ui}$ denotes the $i$-th item purchased by user $u$ in the past. Next, we define a KG as a multi-relational graph $\mathcal{G} = (\mathcal{V}, \mathcal{E})$, where $\mathcal{V}$ is the set of entity vertices and $\mathcal{E} \subset \mathcal{V} \times \mathcal{R} \times \mathcal{V}$ is the set of edges connecting entities with a relation from $\mathcal{R}$. Each item $v$ has its own KG, $g_v$, comprising an entity set $\mathcal{V}_v$ and a relation set $\mathcal{R}_v$ which
% = \{(h,r,t)|h, t \in V, r \in R\}$, in which an (h,r,t) triple expresses a relationship r between the head $h$ and tail $t$ entities and
contain features
%detailing
of
$v$. We devise a set of item-entity alignments $\mathcal{A} = \{(v,e)|v \in \mathcal{I}, e \in \mathcal{V}\}$, where $(v,e)$ indicates that item $v$ is aligned with an entity $e$.

Given a user $u$ and an item $v$ represented by its KG $g_v$, the task is to generate an explanation of natural language sentences $E_{u,v}$ as to why item $v$ was recommended for the user $u$. 
%Here, the natural language explanation $E_{u,v}$ is a verbalization of $g_v$
%\jun{Is it more accurate to say \enquote{a verbalization of a subset of $g_v$}?  It seems to me an explanation is generated from the entire $g_v$, but the entire $g_v$ is not necessarily verbalized.}\ac{as of right now  our ground-truth data is setup so it is a verbalization of $g_v$. However, I have commented out this sentence, since it is more data specific and not problem specific.}, 
%and thus an item-specific explanation grounded on the given structured input. 
As in previous multi-task explainable recommendation models, KnowRec calculates a rating score $r_{u,v}$ that measures $u$'s preference for $v$. 
%By joint training on both the recommendation and explanation task, we can  contexualized the embeddings 
By jointly training on the recommendation and explanation generation, our model can contextualize the embeddings more adequately with training signals from both tasks.

%The NL supervision signals then help contextualize the embeddings used for the recommendation task. \ac{Tried to address here why we co-train.}

\section{Model}\label{sec:model}
% This section describes details of our KnowRec model.  
\cref{fig:model} illustrates our model with the user-item graph constructed through collaborative filtering signals, an encoder, and inference functions for explanation generation and rating prediction.

\subsection{Input}\label{sec:input}
The input of KnowRec comprises a user $u$ represented by the user's purchase history $\{ v_{ui} \}$ and an item $v$ represented by its KG $g_v$, as introduced in \cref{sec:problem}.
Let $v_c$ denote the item currently considered by the system.  The item $v_c$ is aligned with one of the entities through $\mathcal{A}$ and becomes the center node of $g_v$, as shown in Figure \ref{fig:model}.
%, and the KG is formulated as a graph with $v_c$ its center node, shown in \cref{fig:model}.

Because our system leverages a Transformer-based encoder, we first linearize the input into a string.  For the user $u = \{ v_{ui} \}$, we initialize it by mapping each purchased item $v_{ui}$ into tokens of the item's name.  For the item $v$ represented by $g_v$, we decompose $g_v$ into a set of tuples $\{ t_{vj} \}$, where $t_{vj} = (v_c, r_{vj}, n_{vj})$, $n_{vj} \in \mathcal{V}_v$, and $r_{vj} \in \mathcal{R}_v$. We linearize each tuple $t_{vj}$ into a sequence of tokens using lexicalized names of the nodes and the relation.  We then concatenate all the user tokens and the item tokens to form the full input sequence $x$.
%Then, the full input sequence is denoted as $x = [u; v] = [v_{u1}, \dots ,v_{uj}, t_{v1}, \dots, t_{vk}]$.
%linearize $g_v$ into a sequence of tokens $[g_{i1}, g_{i2}, \dots, g_{ik}]$ where 
%, where the currently considered item $v_c$ constitutes the center node of the graph.
%Let $t_{vi}$ denote the $i$-th tuple in $g_v$. $t_{vi} = (v_c, r_{vi}, n_{vi})$. We linearize $t_{vi}$ into a sequence of tokens.
%$x = [v_{u1}, \dots ,v_{un}] + [v_c, r_{i1}, n_{i1}, ..., v_c, r_{ik}, n_{ik}]$
%KnowRec takes as input: query $q$, item $v$, query mask $m_q$, and item mask $m_v$. 
%As we treat $u$ as a user's purchase history, we initialize $u$ as the set of natural language tokens which make up the name of a user's previously purchased items.
%\jun{How is this done specifically? Is it done so by listing names of all products purchased by $u$?} \ac{I tried to make it more clear, by mentioning "names of user's purchased items"}
%Similarly, we initialize an item $v$ as the set of natural language tokens which make up the item's KG. Because our system leverages transformer-based encoders, we first 
%linearize the KG $g_v$ into a string, where the currently considered item $v_c$ constitutes the center node of the graph. 
%Thus, our full input sequence is denoted as $X = [u, v] = [v_{u1}, \dots, v_{un}, g_{i1}, \dots, g_{ik}]$, where $v_{u1}, \dots, v_{un}$ and $g_{i1}, \dots, g_{ik}$ represent a user's purchase history and linearized item KG, where $g_{i1}$ denotes the i-th component in the KG.
For example, suppose the current item $v_c$ is the book \textit{Harry Potter}, the KG has a single tuple (\textit{Harry Potter}, \textit{author}, \textit{J.K. Rowling}), and the user previously purchased two books \textit{The Lord of the Rings} and \textit{The Little Prince}. In this case, input sequence $x =$ \textit{The Lord of the Rings The Little Prince Harry Potter author J.K. Rowling}.
%user $u$ is represented as \textit{\enquote{The Lord of the Rings Little Prince}}, and a tuple $t_{vi}$ could be \textit{\enquote{Harry Potter author J.K. Rowling}}.
%$i$ or [$g_{i1}, \dots, g_{im}$], would be \textit{\enquote{Harry Potter author J.K. Rowling}}, while $u$ or $[v_{u1}, \dots, v_{un}]$ would be \textit{"The Lord of the Rings Little Prince"}.
%Additionally, in our model, we add special denotation tokens <[head], [rel], [tail]> at the start of each graph component, in order to add additional context to the flattened graph string. 

%In order to properly encode these tokens, we can either construct a dictionary, mapping integers to tokens or leverage a pretrained tokenizer such as from BERT~\citep{devlin2018bert}. 
We map the tokens to randomly initialized vectors or pre-trained word embeddings such as those in BART~\cite{lewis-etal-2020-bart},
%We map the tokens to an embedding vector for which we can use a matrix with randomly initialized parameters or pre-trained tokenizer such as BART~\cite{lewis-etal-2020-bart},
obtaining $\mathbf{X}_0 = [\dots ; \mathbf{V}_{ui}; \dots ; \mathbf{T}_{vj}; \dots]$ where $\mathbf{V}_{ui}$ and $\mathbf{T}_{vj}$ are word vector representations of $v_{ui}$ and $t_{vj}$, respectively. Unlike previous work on KG recommendation~\citep{wang2020ckan} where users/items are represented via purchase history and propagated KG information, our system infuses KG components to provide a recommendation and its natural language explanation. Our system also differs from prior studies on explainable recommendation in that while they focus on reviews and thus encode users/items as random vectors with additional review-based sparse token features as auxiliary information~\citep{li-etal-2021-personalized}, we directly encapsulate KG information into the input representation.
%Where previous work \jun{on } has focused on reviews and thus encoded users/items as random vectors with additional review-based sparse token features as auxiliary information~\citep{li-etal-2021-personalized}, we directly encapsulate KG information into $\mathbf{X}_0$.
%\jun{The notation $\mathbf{I}$ has been introduced here, but is it used in any other places?}\ac{Thank you for pointing that out. It should be changed to X. Please review the new changes.}\jun{Okay, I fixed.}
%Finally, masks $m_q$ and $m_v$ denote which tokens belong to the query/item for the recommendation task. 
%When training the system for recommendation explanation, we expose the model to ground truth justifications/explanations from which the KG is expected to verbalize. This process, termed teacher forcing, is a prevalent strategy when training models which expect sequential types of output. 

\subsection{Encoder}\label{sec:encoder}
\textbf{Collaborative KG Representation}. 
%To encode the graph-topology into KnowRec, 
%we adopt a similar query/item representation as in CKAN~\citep{wang2020ckan}, serving as the basis of our rating prediction module. 
%CKAN first represents users  as the vector set of items purchased, and represents items as a vector set of items purchased by the same user. 
%CKAN then propagates this representation through connections in an associative knowledge graph. A knowledge-aware attention mechanism learns the weights of entities in a corresponding entity set to generate a weighted representations of entities which represent a user and item. The graph components may be represented via their tokenization or as unique integers. 
Because KnowRec outputs a natural language explanation grounded on KG facts, as well as a recommendation score for the user-item pair, we need to construct a user-item-linked KG to represent an input through its corresponding lexical graph feature. To do so, we leverage collaborative signals from $\mathbf{Y}$,
%\jun{Is this actually $\mathbf{Y}$ introduced in the beginning of \cref{sec:problem}?}\ac{Yes, please see the first sentence of section 3 where we define the user-item interaction matrix.}\jun{Okay, changed $Y$ to $\mathbf{Y}$ to make it consistent.},
combining $u$ with $v$ by linking previously purchased products $v_{ui}$ to the current item $v_c$ from $g_v$, forming a novel lexical user-item KG. Additionally, we connect all previously purchased items together in order to graphically
model collaborative filtering effects
%encode a user via their purchase history
for rating prediction, as illustrated in Figure~\ref{fig:model}. Note that the relations between previously purchased items and the current items do require a lexical representation for our model. The resulting graph goes through the Transformer architecture, as described below.
%The user-item KG's entity set is then defined as:
%\begin{equation}
%  \mathcal{V}_{uv} = \{e|(v,e) \in \mathcal{A}\ \textrm{and} \  v|y_{uv}=1 \ \textrm{and} \ e \in \mathcal{V}_v\}
%\end{equation}

% \ac{Need some assistance with defining the new KG in set notation}\\

\noindent
\textbf{Global Attention}. 
%As Transformer architectures have seen significant performance gains in numerous natural language generation tasks, they have also recently been adopted for the personalized explainable recommendation task~\citep{li-etal-2021-personalized}.
Transformer architectures have recently been adopted for the personalized explainable recommendation task~\citep{li-etal-2021-personalized}.
%We similarly leverage such mechanism where $\mathbf{X}$ is encoded via an attention mechanism, acting as a global attention.
We similarly leverage Transformer encoder layers~\cite{vaswani2017attention}, referred to as Global Attention, to encode the input representation
%$\mathbf{X}$
with self-attention as:
%Global attention consists of $L$ layers, which encode the previous layer's output of $\mathbf{X}_{l-1}$ to $\mathbf{X}_{l}$.
%\jun{Describing subscripts (indexes) in bold font seems rare. Is it intentional?}.
%At the $l$-th layer, self-attention is defined as: 
\begin{equation}\label{eq:global_att}
  \begin{aligned}
   \mathbf{X}_l &= \operatorname{Attn}(\mathbf{Q}, \mathbf{K}, \mathbf{V})=\operatorname{softmax}\left(\frac{\mathbf{Q} \mathbf{K^{\top}}}{\sqrt{d_{k}}}\right) \mathbf{V}, \\
    \mathbf{Q} &= \mathbf{X}_{l-1} \mathbf{W}_l^Q, \mathbf{K} = \mathbf{X}_{l-1} \mathbf{W}_l^K, \\
    \mathbf{V} &= \mathbf{X}_{l-1} \mathbf{W}_{l}^V 
  \end{aligned}
\end{equation}
where $\mathbf{X}_{l}$
%and $\mathbf{X}_{l-1}$ are the outputs of the current $l$-th layer and previous layer.
is the output of the $l$-th layer in the encoder, and $\mathit{d}_{k}$ is a tunable parameter.
$\mathbf{Q}$, $\mathbf{K}$, and $\mathbf{V}$ represent the \textbf{Q}uery, \textbf{K}ey, and \textbf{V}alue vectors, respectively, each of which is calculated with the corresponding parameter matrix $\mathbf{W}$ in the $l$-th layer.
%Here, $\mathbf{Q}$ represents a \textbf{Q}uery, $\mathbf{K}$ a \textbf{K}ey, and $\mathbf{V}$ a \textbf{V}alue.
%Note, we differentiate query $\mathbf{Q}$ when calculating attention from query $q$ in our input. 
%$\mathbf{W}$ are learnable parameters, with size ${d_k \times d_k}$, where $d_k$ is the dimension of word vectors\zz{please check and modify}.
%\jun{To improve mathematical soundness, please specify the dimension of matrices concisely when they are firstly introduced, e.g., $\mathbf{W} \in \mathbb{R}^{d_k \times d_k}$ is learnable parameters ...} 
%While previous work on explainable recommendation has used a modified transformer encoder, we assume a fully connected encoder as an initial representation of the user and item, updating user $u$ and item $v$ into its initial representation, wherein our next module updates this representation for both the recommendation and explanation task. 
Note that the transformer encoder may be initialized via a pre-trained language model. 

\noindent
\textbf{User-Item Graph Attention}. We further propose User-Item Graph Attention encoder layers, which compute graph-aware attention via a mask to capture the user-item graph's topological information, which runs in parallel with the Global Attention encoder layers. 

We first extract the mask $\mathbf{M}_g \in \mathbb{R}^{m \times m}$ from the user-item linked KG, where $m$ is the number of relevant KG components, i.e., nodes and edges that are lexically expressed in the KG (edges between $v_{ui}$ and $v_c$ not included).  In $\mathbf{M}_g$, each row/column refers to a KG component. $M_{ij} = 0$ if there is a connection between component $i$ and $j$ (e.g., \enquote{J.K. Rowling} and \enquote{author}) and $-\infty$ otherwise. In addition, we assume all item components, i.e., the previous purchases and the current item, are mutually connected when devising $\mathbf{M}_g$. 

%For each $l$-th layer, 
For each layer (referred to as the $l$-th layer),
we then transfer its input $\mathbf{X}_{l-1}$ into a component-wise representation $\mathbf{X}_{l-1}^g \in \mathbb{R}^{m \times d}$, where $d$ is the word embedding size.  Motivated by~\citet{ke-etal-2021-jointgt}, we perform this transfer by employing a pooling layer that averages the vector representations of all the word tokens contained in the corresponding node/edge names per relevant KG component. With the transferred input $\mathbf{X}_{l-1}^g$, we proceed to encode it using User-Item Graph Attention with the graph-topology-sensitive mask as follows:
\begin{equation}\label{eq:graph_att}
  \begin{aligned}
\mathbf{\Tilde{X}}_l^g
  &= \operatorname{Attn}_{M}(\mathbf{Q'}, \mathbf{K'}, \mathbf{V'}) \\
  &= \operatorname{softmax}\left(\frac{\mathbf{Q' K'^{\top}}}{\sqrt{d_{k}}} + \mathbf{M}_g \right) \mathbf{V'}.
  \end{aligned}
 \end{equation}
%\begin{align}\label{eq:graph_att_old}
%\mathbf{\Tilde{X}}_l^g
%  &= \operatorname{Attn}_{M}(\mathbf{Q'}, \mathbf{K'}, \mathbf{V'}) \\ \nonumber
%  &= \operatorname{softmax}\left(\frac{\mathbf{Q' K'^{\top}}}{\sqrt{d_{k}}} + \mathbf{M}_g \right) \mathbf{V'}.
%\end{align}
%\begin{equation}
%\begin{split}
%    \mathbf{\Tilde{X}}_l^g = \operatorname{Attn}_{M}(\mathbf{Q'}, \mathbf{K'}, \mathbf{V'})= \\ 
%    \operatorname{softmax}\left(\frac{\mathbf{Q' K'^{\top}}}{\sqrt{d_{k}}} + \mathbf{M}_g \right) \mathbf{V'}.
%\end{split}
%\end{equation}
where query $\mathbf{Q'}$, key $\mathbf{K'}$, and value $\mathbf{V'}$ are computed with the %transferred
transferred input and
%corresponding
learnable parameters in the same manner as \cref{eq:global_att}.

Lastly, we
%proceed to\jun{This sounds a little redundant, as we used it above.}
combine the outputs of the Global Attention encoder and the User-Item Graph Attention encoder in each layer. 
As the two outputs have different dimensions, we first expand $\mathbf{\Tilde{X}}_l^g$ to the same dimension of $\mathbf{X}_l$ through a \textit{gather} operation, i.e., broadcasting each KG component-wise representation in $\mathbf{\Tilde{X}}_l^g$ to every encompassing word of the corresponding component and connecting those representations. 
%As the two outputs have different dimensions, we first \textit{gather} the word representations from the graph components in $\mathbf{\Tilde{X}}_l^g$ by expanding $\mathbf{\Tilde{X}}_l^g$ to the same dimension $\mathbf{\Tilde{X}}_l^g$, broadcasting the component representation in $\mathbf{\Tilde{X}}_l^g$ to each of its encompassing words to generate a vector representation with the same dimension as $\mathbf{X}_l$. 
%\zz{it will be nicer to modify this sentence to briefly explain what "gather" actually does in this case here. Maybe something like "repeating the same component representation in $\mathbf{\Tilde{X}}_l^g$ for each word in a component to generate a vector representation with the same dimension as X-l"? }\ac{Thank you for the advice. I have incorporated it with a couple of changes. Please feel free to trim it if it is too long.}
We then add the expanded $\mathbf{\Tilde{X}}_l^g$ to $\mathbf{X}_l$ through element-wise addition, generating the $l$-th encoding layer's output:

% global encoding $\mathbf{X}_l$ which captures attention between different tokens to the user-item KG encoding $\mathbf{\Tilde{X}}_l^g$, we \textit{gather} the word representations from the graph components in $\mathbf{\Tilde{X}}_l^g$ by expanding $\mathbf{\Tilde{X}}_l^g$ to the same dimension $\mathbf{\Tilde{X}}_l^g$ and add the two representations through element-wise addition, generating the l-th encoding layer's output:
%\zz{how about change this paragraph into two sentences: "We further combine the outputs of the Global Attention encoder and the User-Item Graph Attention encoder of the same layer. As the two outputs have different dimensions, we first gather the word representations..., and then add the two..., generating a final l-th encoding layer's output:"}\ac{If we make it into 2 sentences, do you think its worth putting as a separate paragraph?}\zz{yes, see my previous comment, it should be a separate paragraph.}\ac{Thanks! Done.}
%\zz{should 'l' be 'L' in (4) to be consistent with Figure 2? by the final encoding layer, do you mean the Output Module in Fig. 2? PS, it is better to say "...to generate a hidden representation to feed into..."}\ac{After reviewing our code again I have changed the Figure 2 as this happens in each l-th layer. Thank you for catching this.}
\begin{equation}
    \mathbf{\Tilde{X}}_l = gather(\mathbf{\Tilde{X}}_l^g) + \mathbf{X}_l
\end{equation}

Note, in this section, we illustrate the Global Attention encoder, User-Item Attention encoder, and their combination with single-head attention. In practice, we implement both encoders with multi-head attention as in~\citet{vaswani2017attention}.
%\zz{I just modified this sentence a bit. Please check.}
%Note, in both attention encoders we describe our model with a single-head attention, but in practice implement multi-head attention.
\subsection{Rating Prediction}
For the rating prediction task, we first separate and isolate user $u$ and item $v$ features via masking. Once isolated, we perform a mean pool on all their respective tokens and linearly project $u$ and $v$ to perform a dot-product between the two new vector representations as follows:
\begin{equation}\label{eq:rating}
  \begin{aligned}
   \mathbf{\Tilde{x}}_u &= pool_{mean}(\mathbf{\Tilde{X}}_L + \mathbf{m}_u)\mathbf{W}^u\\
   \mathbf{\Tilde{x}}_v &= pool_{mean}(\mathbf{\Tilde{X}}_L + \mathbf{m}_v)\mathbf{W}^v\\
   \hat{r}_{u,v} &= dot(\mathbf{\Tilde{x}}_u, \mathbf{\Tilde{x}}_v),
  \end{aligned}
\end{equation}
where $\mathbf{m}_u$ and $\mathbf{m}_v$ are the user and item masks that denote which tokens belong to the user and item, $\mathbf{W}$s are learnable parameters, and \textit{L} refers to the last layer of the encoder. %of size $d_k \times d_k$\jun{We may need to change $d_k$}. 

\subsection{Explanation Generation}
Before generating a final output text for our explanation, we pass the representation through a fully connected linear layer as the encoder hidden state and decode the representation into its respective output tokens through an auto-regressive decoder, following previous work~\cite{lewis-etal-2020-bart}.

% \jun{I mostly copied and pasted relevant contents into \cref{sec:model}, but am not sure which contents should be put into \cref{sec:generation,sec:recommendation}. @Anthony, can you revise the description?}

% \subsection{Explanation Generation}\label{sec:generation}

% \subsection{Recommendation Scoring}\label{sec:recommendation}

\subsection{Joint-learning Objective}\label{sec:train}
% Each explanation in the dataset is dependent on its respective item. Thus, to avoid data leaking into the train/development/test set, we divide the train/development/test by item. That is, all items in the test set are unseen during training. \zz{This paragraph should be put in the experiments part}\ac{Actually, I had it there and will delete it from this section. Thank you.}
%The train set can be seen as older items in the catalogue that have a history of purchase. 
%During training time we are given the ground truth item explanations to perform teacher forcing on the model, where the ground truth is given as subsequent input instead of the generated output of previous tokens.

%Our input/output training setup then consists of:
% In the training phase, the input of our system consists of:
% %\begin{itemize}
% %\item Input
% \begin{itemize}
%     \item Past query-item associations (e.g., user-item matrix, purchase history)
%     \item Item descriptions
%     \item Item knowledge graph
% \end{itemize}
% %\item Output
% The output of our system comprises:
% \begin{itemize}
%   \item Explanation% (as a subset of item descriptions) $\rightarrow$ gold standard
%   \item Rating prediction score
% \end{itemize}
% %\end{itemize}
% Explanations $E$ are extracted from item descriptions $D$: $E \subseteq D$.

As previously noted, our system consists of two outputs: a rating prediction score $\hat{r}_{u,v}$ and natural language explanation $E_{u,v}$ which justifies the rating by verbalizing the item's corresponding KG. We thus perform multi-task learning to learn both tasks and manually define regularization weights $\lambda$, as in similar multi-task paradigms, to weight the two tasks. Taking  $\mathcal{L}_r$ and $\mathcal{L}_e$ to represent the recommendation and explanation cost functions, respectively, the multi-task cost $\mathcal{L}$ then becomes: 
\begin{equation}
    \mathcal{L} = \lambda_r\mathcal{L}_r + \lambda_e\mathcal{L}_e,
\end{equation}
where $\lambda_r$ and $\lambda_e$ denote the rating prediction and explanation regularization weights, respectively.

We define $\mathcal{L}_r$ using Mean Square Error (MSE) in line with conventional item recommendation and review-based explainable systems:
\begin{equation}
    \mathcal{L}_r = \frac{1}{|\mathcal{U}| |\mathcal{I}|}\sum_{u \in \mathcal{U} \wedge v \in \mathcal{I}}({r}_{u,v}-\hat{r}_{u,v})^2,
\end{equation}
where %$(U,V)$ denotes the total set of user-item pairs and
$r_{u,v}$ denotes the ground-true score.

Next, as in other NLG tasks~\citep{lewis-etal-2020-bart, zhang2020explainable}, we incorporate Negative Log-Likelihood (NLL) as the explanation's cost function $\mathcal{L}_e$. Thus, we define $\mathcal{L}_e$ as:
%\zz{Some notations in (8) need to be explained.}\ac{Added more below. Please check.}
\begin{equation}
    \mathcal{L}_e = \frac{1}{|\mathcal{U}| |\mathcal{I}|}\sum_{u \in \mathcal{U} \wedge v \in \mathcal{I}}\frac{1}{|E_{u,v}|}\sum_{t=1}^{|E_{u,v}|}-\log {p}_t^{e_t}
\end{equation}
where ${p}_t^{e_t}$ is the probability of a decoded token $e^t$ at time step t.

\section{Dataset}\label{sec:dataset}
% \ac{Should we just call this section "Datasets"}
Although KG-recommendation datasets exist, they do not contain any supervision signals to NL descriptions. Thus, to evaluate our explainable recommendation approach in a KG-aware setting and our KnowRec model, we introduce two new datasets based on the Amazon-Book and Amazon-Movie datasets~\cite{he2016ups}: (1) Book KG-Exp and (2) Movie KG-Exp.

Recall that our task requires an input KG along with an NL explanation and recommendation score. Because it is more efficient to extract KGs from text, rather than manually annotate each KG with text, we take a description-first approach, automatically extracting KG elements from the corresponding text. Given the currently available data, we leverage item descriptions as a proxy for the NL explanations, while constructing a user-item KG from an item's features and user's purchase history.

We first extract entities from a given item description via DBpedia Spotlight~\cite{mendes2011dbpedia}, a tool that detects mentions of DBpedia~\cite{auer2007dbpedia} entities from NL text. We then query for each entity's most specific type and use those types as relations that connect the item to its corresponding entities. We construct a user KG via their purchase history, e.g. $[Purchase_1, Purchase_2,... Purchase_n]$, as a complete graph where each purchase is connected. Finally, we connect all the nodes of the user KG to the item KG, treating each user purchase as a one-hop neighbor of the current item. To ensure the KG-explanation correspondence, we filter out any sentences in the explanation in which no entities were found. To measure objectivity, we calculate the proportion of a given KG's entities that appear in the explanation, called entity coverage (EC) (defined in~\cref{sec:entity-coverage}). We summarize our dataset statistics in \cref{tab:data-summary} and present a more comprehensive comparison in Appendix~\ref{sec:dataset-comparison}.

\begin{table*}[!ht]
\centering
\small
% \resizebox{\textwidth}{!}{%
\setlength\tabcolsep{5pt}
\begin{tabular}{%r|
l@{\smallcol}r@{\smallcol}r@{\smallcol}r@{\smallcol}c@{\smallcol}r@{\smallcol}r@{\smallcol}r@{\smallcol}r@{\smallcol}c@{\smallcol}r
%l|lllllllll
}
\toprule
%\hline
%Task &
  \multicolumn{1}{c}{Name} &
  \multicolumn{1}{c}{\#Users} &
  \multicolumn{1}{c}{\#Items} &
  \multicolumn{1}{c}{\#Interactions} &
  \multicolumn{1}{c}{KG} &
  \multicolumn{1}{c}{\#Es} &
  \multicolumn{1}{c}{\#Rs} &
  \multicolumn{1}{c}{\#Triples} &
  \multicolumn{1}{c}{EC} &
  \multicolumn{1}{c}{Desc.} &
  %\begin{tabular}[c]{@{}l@{}}Words/\\Sample \end{tabular} \\
  Words/Sample \\
\midrule
%\hline
%KG-Rec  & 
% Last.FM             & 23,566  & 48,123  & 3,034,796  & Yes & 58,266  & 9  & 464,567   & No & -    \\
% %KG-Rec  & 
% Book-Crossing       & 276,271 & 271,379 & 1,048,575  & Yes & 25,787  & 18 & 60,787    & No & -    \\
% %KG-Rec  & 
% Movie-Lens20M       & 138,159 & 16,954  & 13,501,622 & Yes & 102,569 & 32 & 499,474   & No & -    \\
% %KG-Rec  & 
% Amazon-book (KG)    & 70,679  & 24,915  & 847,733    & Yes & 88,572  & 39 & 2,557,746 & No & -    \\ %\midrule
% \hline
% %Exp-Rec & 
% Yelp-Restaurant     & 27,147  & 20,266  & 1,293,247  & No  & -      & - & -        & No & 12.32 \\
% %Exp-Rec & 
% Amazon Movies & 7,506   & 7,360   & 441,783    & No  & -      & - & -        & No & 14.14 \\
% %Exp-Rec & 
% TripAdvisor-Hotel   & 9,765   & 6,280   & 320,023    & No  & -      & - & -        & No & 13.01 \\ 
% %\midrule
% \hline
%\textit{KG-Exp} &
  \textit{Book KG-Exp} &
  \textit{396,114} &
  \textit{95,733} &
  \textit{2,318,107} &
  \textit{Yes} &
  \textit{195,110} &
  \textit{392} &
  \textit{745,699} &
  \textit{71.45} &
  \textit{Yes} &
  \textit{99.96} \\
%\textit{KG-Exp} &
  \textit{Movie KG-Exp} &
  \textit{131,375} &
  \textit{18,107} &
  \textit{788,957} &
  \textit{Yes} &
  \textit{59,036} &
  \textit{363} &
  \textit{146,772} &
  \textit{71.32} &
  \textit{Yes} &
  \textit{96.35} \\
\bottomrule
%\hline
\end{tabular}%
% }
\caption{\label{tab:data-summary}Statistics of our Book KG-Exp and Movie KG-Exp benchmark datasets.%\jun{Please explicitly explain which of KG-REC, Exp-Rec, and KG-Exp corresponds to which of the three categories separated by horizontal lines.}\ac{added context to the caption above}\jun{Looks good!}
\textit{\#Es}, \textit{\#Rs}, and \textit{Desc.} denote number of entities, number of relations, and if the dataset contains parallel descriptions. 
% We refer to the proposed book and movie datasets as \textit{Book KG-Exp} and \textit{Movie KG-Exp}, respectively.
%\jun{I think a dataset's name is good if it is as short as possible and clear about its origin (if any). Given that, I wonder if \enquote{AMZ-Book KG-Exp} or more simply \enquote{Book KG-Exp} is slightly better. What do you think?}\ac{I like Book KG-Exp and will change it. Thank you.}
}
% \vspace{-0.4cm}
\end{table*}

\section{Experiments}\label{sec:experiments}
\begin{table*}[h]
\small
\centering
% \resizebox{\textwidth}{!}{%
\begin{tabular}{@{\smallcol}p{1.6cm}@{\smallcol}l@{\smallcol}r@{\smallcol}r@{\smallcol}r@{\smallcol}r@{\smallcol}r@{\smallcol}r@{\smallcol}r@{\smallcol}r@{\smallcol}r@{\smallcol}r@{\smallcol}}
\toprule
Dataset & Model & BLEU-1 & BLEU-4 & USR & R2-F  & R2-R  & R2-P  & RL-F  & RL-R  & RL-P  & EC\\
\midrule
  & Att2Seq  & 8.86 & 0.39    & 0.30  & 2.08    & 1.41    & 8.47    & 8.07    & 11.65    & 9.49   & 0.44\\
  & NRT         & 11.76 & 0.57    & 0.03  & 1.50    & 1.40    & 3.25    & 7.20    & 11.70    & 8.05   & 0.98\\
Movie & Transformer & 8.67 & 0.18    & 0.33  & 1.21    & 0.91    & 6.55   & 6.58    & 9.54   & 9.69  & 0.82  \\
KG-Exp & PETER       & 14.66 & 3.99    & 0.55  & 5.07    & 4.26    & 11.66   & 15.06   & 16.67   & 23.03 & 10.58   \\
  & PEPLER      & 11.68 & 0.13  & 0.46 & 0.56  & 0.63  & 0.54  & 8.90  & 10.92   & 9.53 & 0.78 \\
\cline{2-12}
  & KnowRec     & \textbf{37.02} & \textbf{10.71}  & \textbf{0.83} & \textbf{15.49} & \textbf{15.12} & \textbf{18.15} & \textbf{27.71} & \textbf{28.71} & \textbf{37.10} & \textbf{67.97}\\
\midrule
 & Att2Seq     & 19.51 & 1.85  & 0.43 & 5.08  & 3.76  & 12.15 & 12.98 & 16.55 & 20.89   & 0.86\\
 & NRT         & 21.06 & 2.59  & 0.10 & 6.18  & 4.88  & 11.44  & 15.57 & 18.67 & 24.36   & 1.57\\
Book & Transformer & 16.90 & 2.01  & 0.12 & 5.68  & 4.23  & 11.94  & 13.66 & 15.57 & 26.87   & 2.08\\
KG-Exp & PETER       & 27.93 & 8.39  & 0.71 & 11.94 & 10.36 & 18.68 & 21.24 & 23.30 & 28.02  & 17.39\\
 & PEPLER      & 16.07 & 1.20  & 0.90 & 2.39  & 2.63  & 2.26  & 13.03 & 16.34 & 12.24  & 0.74\\
\cline{2-12}
 & KnowRec     & \textbf{38.53} & \textbf{12.60} & \textbf{0.92} & \textbf{19.78} & \textbf{19.44} & \textbf{23.22} & \textbf{28.29} & \textbf{29.43} & \textbf{35.28} & \textbf{69.50}\\
\bottomrule
\end{tabular}%
\caption{\label{tab:fullresults}Comparison of neural generation models on the Movie KG-Exp and Book KG-Exp datasets.%\jun{Here I have updated \cref{tab:fullresults_bkup} mainly for conciseness. Please update \cref{tab:fewshot-reusults} accordingly.}
%\jun{If this last sentence also appears in the caption of \cref{tab:fewshot-reusults}, it is redundant, and we should move it to the body text to save space.}\ac{Actually, I did mention it in the body text, but had forgotten. Thank you for pointing this out and for the nice looking table.}
%Note, we test on a sample of the full test set for efficiency due to computational limitations.
}
\end{table*}

\subsection{Evaluation Metrics}
We assess explainable recommendation following prior work: 1) on the recommendation performance and 2) on the explanation performance. For the explanation generation task, we employ standard natural language generation (NLG) metrics: BLEU~\cite{papineni-etal-2002-bleu} and ROUGE~\cite{lin-2004-rouge}. We measure diversity and the detail-oriented features of the generated sentences using Unique Sentence Ratio (USR)~\cite{li2020generate, li-etal-2021-personalized}, and use EC, instead of feature coverage ratio, for coverage due to our non-review-based explanations.

\subsection{Implementation}
We train our newly proposed KnowRec model on two settings of the Book and Movie KG-Exp datasets, a full training set and a few-shot setting, where 1\% of the data is used. Because our method provides item-level explanations based on KGs, we split the datasets based on their labeled description/explanation, and as such, we experiment in a setting where items in the test set can be unseen during training. By doing so, we handle a unique case that has not been considered in previous research relying on item reviews. The train/validation/test sets are split into 60/20/20. For KnowRec, we use BART as our pre-trained model, with a Byte-Pair Encoding (BPE) vocabulary~\cite{radford2019language}. We compare our approach to available explanation generation baselines, including those that leverage user and item IDs, and those which utilize word-level features. We adapt the baselines to use the KG information and detail them in Appendix~\ref{sec:baselines}.
% Because of computation limitations, for evaluation purposes, we randomly sample and evaluate on 1\% of the test set, containing 4,491 and 1,456 samples for the Book and Movie dataset respectively. Note, that the size of the test set
% %\jun{What do you mean by the pronoun \enquote{this}? Does it mean the size of test data?}\ac{Please see change. Thank you.}
% is comparative to other text generative tasks such as KG-to-text~\cite{gardent-etal-2017-webnlg} and summarization~\cite{nallapati-etal-2016-abstractive}. 
For more details regarding our experimental settings please see Appendix~\ref{sec:hyper-settings}.

\begin{table*}
\small
\centering

\begin{tabular}{@{\smallcol}p{1.6cm}@{\smallcol}l@{\smallcol}r@{\smallcol}r@{\smallcol}r@{\smallcol}r@{\smallcol}r@{\smallcol}r@{\smallcol}r@{\smallcol}r@{\smallcol}r@{\smallcol}r@{\smallcol}}
\toprule
Dataset & Model & BLEU-1 & BLEU-4 & USR & R2-F  & R2-R  & R2-P  & RL-F  & RL-R  & RL-P & EC \\
\midrule
  & Att2Seq  & 2.63 & 0.00  & 0.00  & 0.00  & 0.00  & 0.00  & 2.73  & 4.25  & 2.63 & 0.01    \\
Movie  & NRT  &  8.78 & 0.32   & 0.01 & 1.84  & 1.08  & 11.73 & 7.12  & 10.17 & 17.97   & 0.07\\
KG-Exp & Transformer & 12.23 & 0.27   & 0.16  & 1.24  & 1.07  & 3.54  & 6.97  & 9.54  & 12.00   & 1.18\\
(Few-shot) & PETER & 12.28 & 0.68   & 0.36  & 2.33  & 1.45  & 12.49 & 12.00 & 13.18 & 18.03 & 5.44\\
  & PEPLER &  12.58 & 0.41   & 0.01 & 1.26  & 1.44  & 1.18  & 10.73 & 12.63 & 10.38 & 0.11\\
\cline{2-12}
  & KnowRec   &  \textbf{33.89} & \textbf{7.53}  & \textbf{0.87} &\textbf{13.41} & \textbf{12.60} & \textbf{17.67} & \textbf{24.48} & \textbf{25.63} & \textbf{35.66} & \textbf{63.92} \\
\midrule
 & Att2Seq    &   16.58 & 1.53 & 0.22 & 4.68  & 3.10  & 15.58 & 13.30 & 15.28 & 21.32 & 0.26\\
Book & NRT          &   19.12 & 2.19 & 0.01 & 6.11  & 4.36  & 13.99 & 15.18 & 20.47 & 16.78 & 1.19\\
KG-Exp & Transformer &  12.69 & 1.22 & 0.08 & 3.60  & 3.16  & 8.65  & 9.77  & 15.64 & 10.58 & 1.57\\
(Few-shot) & PETER       &   18.38 & 2.87 & 0.45 & 7.12  & 5.07  & 17.50 & 14.74 & 17.66 & 17.52 & 4.23\\
 & PEPLER      &   7.96 & 0.26 & 0.02 & 0.67  & 0.63  & 0.83  & 7.59  & 10.07 & 7.04 & 0.54\\
\cline{2-12}
 & KnowRec    & \textbf{28.93} & \textbf{7.94} & \textbf{0.93} & \textbf{17.28} & \textbf{16.05} & \textbf{22.45} & \textbf{24.84} & \textbf{25.19} & \textbf{36.60} & \textbf{60.46} \\
\bottomrule
\end{tabular}%
\caption{\label{tab:fewshot-results}Comparison of neural generation models on the Movie KG-Exp and Book KG-Exp datasets in the few-shot learning setting (1\% of training data).}
% }
\end{table*}

\begin{table}[]
\centering
\def\collen{0.6cm}
\scriptsize
%\resizebox{\columnwidth}{!}{%
\begin{tabular}{@{}l|@{\tinycol}p{\collen}@{\tinycol}p{\collen}@{\tinycol}p{\collen}@{\tinycol}p{0.5cm}|@{\tinycol}p{\collen}@{\tinycol}p{\collen}@{\tinycol}p{\collen}@{\tinycol}p{\collen}@{}}
\toprule
%\cmidrule(l){2-9}
% & \multicolumn{2}{c}{Book} &
%  \multicolumn{2}{c|}{\begin{tabular}[c]{@{}c@{}}Book\\ (Few)\end{tabular}} &
%  \multicolumn{2}{c|}{Movie} &
%  \multicolumn{2}{c}{\begin{tabular}[c]{@{}c@{}}Movie\\ (Few)\end{tabular}} \\ \cmidrule(l){2-9} 
  & \multicolumn{4}{c|@{\tinycol}}{Book KG-Exp} & \multicolumn{4}{c}{Movie KG-Exp} \\
\multicolumn{1}{c|@{\tinycol}}{Model}  & \multicolumn{2}{c@{\tinycol}}{All} & \multicolumn{2}{c|@{\tinycol}}{Few} & \multicolumn{2}{c}{All} & \multicolumn{2}{c}{Few} \\
  & R    & M    & R    & M    & R    & M    & R    & M  \\
\midrule
PMF     & 3.50 & 3.35 & 3.50 & 3.35 & 3.31 & 3.08 & 3.32 & 3.08 \\
SVD++   & 1.03 & 0.80 & 1.01 & 0.64 & 1.20 & \textbf{0.79} & 1.25 & 0.98 \\
NRT     & 0.98 & 0.74 & 1.07 & 0.73 & 1.17 & 0.93 & 1.23 & 0.97 \\
PETER   & 1.01 & 0.79 & 1.03 & 0.82 & 1.24 & 1.03 & 1.24 & 1.00 \\
PEPLER  & \textbf{0.96} & \textbf{0.72} & 1.07 & 0.72 & \textbf{1.14} & 0.91 & 1.27 & 0.96 \\
\hline
%\midrule
KnowRec & 1.04 & 0.75 & 1.04 & 0.72 & 1.22 & 0.92 & \textbf{1.21} & \textbf{0.93} \\
\bottomrule
\end{tabular}%
%}
\caption{Performance comparison on the recommendation task with respect to RMSE and MAE, denoted as R and M on the table respectively.}
\label{tab:recommendation}
\end{table}
%Address jump in performance better:
% - put reasons why other models are bad for their own task
% - PETER vs ours and PEPLER vs ours

\section{Results and Analysis}\label{sec:analysis}
\subsection{Explanation Results}
In~\cref{tab:fullresults}, we evaluate the models' text reproduction performance using BLEU and ROUGE metrics, while also examining their \textit{explainability} through USR and EC analysis.

For BLEU and ROUGE, KnowRec consistently outperforms all baselines, achieving a BLEU-4 score of 10.71 and ROUGE-L F1 score of 27.71 on Movie KG-Exp and a BLEU-4 score of 12.60 and ROUGE-L F score of 28.29 on Movie KG-Exp. This suggests that previous baselines, designed for review-level explanation, are inadequate to produce longer and more objective explanations. Specifically, of the baselines, PETER which utilizes the whole lexical input, adapts best. However, KnowRec makes use of user-item graph encodings, which may lead to better generation of the item KG features mentioned in the ground truth texts. 
% Similarly, we outperform PEPLER, which also leverages a pre-trained language model, but does not adapt well to the KG explanation generation task, diverging from its performance on review explanation~\cite{li2022personalized}. 
While PEPLER \cite{li2022personalized}'s pretrained approach aids in fluent sentence generation, KnowRec excels in generating contextually relevant words around feature-level terms. Unlike PEPLER, which creates concise reviews based on user-item IDs, KnowRec utilizes graph attention to interconnect related components for comprehensive NL text explanations.

In terms of explainability, KnowRec also generates much more diverse sentences (USR), especially compared to models that do not leverage pre-trained models. Note that while PEPLER has a comparable USR score to KnowRec on the Book KG-Exp dataset, it does not similarly produce high-quality and related sentences according to the NLG metrics. Our results show that while the ground truth is based on item-level features, the generated output is still personalized as further discussed in Section~\ref{sec:qual}. Also note the high discrepancy in EC, where the entity-level features are generated in the output text. As our goal is to generate objective and specific explanations, the EC can help real-world users understand what a certain recommended product is about and how it compares to other products. Therefore, it is crucial that explainable models capture these features when producing justifications for recommendations.

\begin{table}[]
\centering
\small
%\resizebox{\columnwidth}{!}{%
\begin{tabular}{@{}l@{\tinycol}l@{\tinycol}l@{\tinycol}l@{\tinycol}|l@{\tinycol}l@{}}
\toprule
              & BLEU-4$\uparrow$ & USR$\uparrow$ & RL-F$\uparrow$  & RMSE$\downarrow$ & MAE$\downarrow$  \\ \midrule
KnowRec       & 7.94   & \textbf{0.93} & 24.84 & \textbf{1.04} & \textbf{0.78} \\
- Recomm.    & \textbf{8.32}   & \textbf{0.93} & \textbf{24.90} & -    & -    \\
- UIG Att. & 7.75   & 0.91 &  24.80 & 1.03 & \textbf{0.78} \\
%No Ratings    & 8.03   & \textbf{0.92} & 17.53 & 25.01 & -    & -    \\
%No User-Item Graph Att. & 7.55   & 0.91 & 17.59 & 24.72 & \textbf{1.03} & \textbf{0.77} \\
%\midrule
\bottomrule
\end{tabular}%
%}
\caption{\label{tab:ablation}Ablation study on the Book KG-Exp (Few-Shot) dataset. \sgquote{Recomm.} means the joint learning with recommendation scoring, and \sgquote{UIG Att.} denotes the user-item graph attention.}
\end{table}

% \begin{table*}[]
% \centering
% \resizebox{\textwidth}{!}{%
% \begin{tabular}{@{}l|ll@{}}
% \toprule
% \multirow{2}{*}{\begin{tabular}[c]{@{}l@{}}Book\\ KG-Exp\end{tabular}} &
%   GT: &
%   \textbf{dan simmons} is the award - winning author of several novels , including the \textbf{new york} times bestsellers \textbf{olympos} and \textbf{the terror} . he lives in \textbf{colorado} . \\
%  & Out: & \textbf{dan simmons} is the author of the \textbf{new york} times bestsellers \textbf{the terror} and the \textbf{olympos} . he lives in colorado springs , \textbf{colorado} . \\ \midrule
% \multirow{2}{*}{\begin{tabular}[c]{@{}l@{}}Movie\\ KG-Exp\end{tabular}} &
%   GT: &
%   \textbf{jules verne} ' s professor lindenbrook leads a trip through monsters , mushrooms and a \textbf{magnetic storm} . \\
%  & Out: & a group of scientists , inspired by \textbf{jules verne} ' s classic novel , take a trip to the \textbf{magnetic storm} at \textbf{the center of the earth} . \\
% \bottomrule
% %\cmidrule(l){2-3} 
% \end{tabular}%
% }
% \caption{Case study of two examples generated by KnowRec from the Book KG-Exp and Movie KG-Exp datasets. GT denotes Ground-Truth. Bold tokens are entities from the user-item KG.}
% \label{tab:case-study}
% \end{table*}

\subsection{Few-shot Explanation Results}
%TODO: rewrite for training data with few item descriptions but item database exists
Real-world recommendation systems may face low-resource problems, where only a small amount of training data with few item descriptions is available but an item database exists.  To reflect this practical situation, we also evaluate a few-shot setting where the training data is 1\% of its total size.
%To test our KG-based approach on the case where the training data contains few item descriptions but an item database exists, we evaluate on a few-shot setting as seen in
%\jun{I think we should start with an explanation on our motivation of this experiment, i.e., why we are interested in this low-resource setting.}\ac{Please see motivation "To test our KG-based approach "}
%Table~\ref{tab:fewshot-results}, where the training data is 1\% of its total size.
As in previous experiments, we set the user-item size for KnowRec to 5. We show the results of this few-shot experiment in \cref{tab:fewshot-results}.  KnowRec consistently and significantly outperforms other explainable baselines on both the Book and Movie datasets in terms of text quality, sentence diversity (USR), and entity representation (ER), showing our approach is effective even in data-scarce scenarios. Like KnowRec, PEPLER also leverages a pre-trained model, namely GPT-2. However, unlike KnowRec, the model does not adapt well to generating item-specific explanations. The second best model, PETER, fully leverages the KG features in their approach. However, such a model does produce diverse sentences. Note that those models that completely rely on user and item IDs, fail to produce quality explanations, as noted by their respective BLEU and ROUGE scores, showing the task to be more complex than previous explanation tasks relying on repetitive, short, and already existing user reviews.

\subsection{Recommendation Performance}
Table~\ref{tab:recommendation} shows the recommendation performance on all KG Explanation datasets.  We report the Root Mean Square Error (RMSE) and Mean Absolute Error (MAE) metrics to evaluate the recommendation task. As shown, all results except PMF are relatively close. PMF significantly underperforms due to the cold start problem presented on new items.
KnowRec achieves performance comparable to other strong baselines, despite KnowRec being the only model that uses lexical features for the recommendation task, while the other models learn the task through user/item IDs. Thus, KnowRec may need more data to learn these parameters. Additionally, because we learn the recommendation task through lexical features, our model provides an interpretable solution that could be directly compared to the produced NL explanations.
%
%hile other baselines slightly outperform KnowRec, recall that KnowRec is the only model which uses lexical features for the recommendation task, while the other models learn the task through user/item IDs. Thus, KnowRec may need more data to learn these parameters. Additionally, because we learn the recommendation task through lexical features, are model provides an interpretable solution which could be directly compared to the produced NL explanations.
%
%all results, except PMF are relatively close, due to the cold start problem presented on new items.\jun{Add an explanation on why PMF is much worse than the other baselines.}\ac{Please see new edits. Thank you.}

\subsection{Ablation Study}
We perform ablation studies to analyze the effects of the recommendation and user-item graph components on Book KG-exp as shown in \cref{tab:ablation}. Due to computational resources, we performed the study on the few-shot setting. We first examine the results of KnowRec without the recommendation module in the second row (\textit{- Recomm.}). By removing the ‘Recomm’ component, the performance on the NLG metrics improves, as the task is now a single-objective generative task instead of a multi-objective. We next study the effects of the User-Item Attention encoders on KnowRec's explainability and recommendation performance (\textit{- UIG Att}).
% \ac{Rewrote this to address 2 reviewer comments: 1. Why few-shot in ablation study? 2. Did we see how our system performs without the recommendation module?}
%\zz{should be "without the User-Item Attention encoders".Note the input to Global Attention is also derived from the User-Item KG}\ac{Thank you for making it clearer.}. 
As shown by \textit{- UIG Att.}, even with a smaller training dataset of 1\% of the full data, by removing this component, we observe a slight decrease in the NLG metrics, BLEU and ROUGE, and less diverse sentences (USR). The representation and attention masking on the user-item graph, which connects and encodes the local item information, may therefore give a better representation of the input which is in turn decoded to produce an explanation. This may be further expressed within larger datasets. Furthermore, from the NLG metric results, we can infer from Table~\ref{tab:ablation} that our rating module does not significantly hinder the performance of the generation component of KnowRec.

\subsection{Qualitative Analysis}
\label{sec:qual}
%objectiveness
%robustness
To grasp KnowRec's effectiveness, we analyze explanations from Movie/Book KG-Exp test sets. These explanations are both grammatically smooth and adept at (1) integrating robust item features for factual insights and (2) tailoring personalized content based on diverse user purchase histories (examples in \cref{sec:examples}, \cref{tab:examples}).

% \zz{Note2: We may need to find a place to talk about the hallucination problem, maybe in Appendix together with the example table or in Limitation part. We may state that hallucination is common to SOTA NLG, and KnowRec still has this problem but it may relieve the problem by leveraging knowledge graph in generation.}

%Taking the first sample of the table pertaining to the movie \textit{Journey to the Center of the Earth},
Consider the first two rows of the table, pertaining to the movie \textit{Journey to the Center of the Earth}.
We can see two different (but syntactically similar) generated explanations for two different users. In one case, the user has bought mystery and fantasy movies such as \textit{Stitch in Crime}, \textit{Columbo}, and \textit{The Lord of the Rings}, and the output integrates related words such as \textit{investigates} and  \textit{mysterious} to personalize the explanation. The second case mentions \textit{classic} and \textit{novel}, possibly because the second user's purchase history involves \textit{Disney} classics and movies based on novels such as \textit{The Hardy Boys} and \textit{Old Yeller}. While the input KG does not explicitly state that \textit{Journey to the Center of the Earth} is a novel, such information may be inferred from the KG's relation and supported through the user's related purchases. In both cases the output closely matches the ground truth, verbalizing item features from the KB such as \textit{Jules Verne} and \textit{magnetic storm}, suggesting that our model is robust in describing the explanation content, while still implicitly reflecting the user's purchase history. 
%book vs movie description. for example the book kgs are typically author related and thus capture more information than the movie data

% We now showcase two examples produced by KnowRec on the Book and Movie datasets, respectively, in Table ~\ref{tab:case-study}.
% %\zz{"on the Book and Movie datasets, respectively, in Table~\ref{tab:case-study}"  }\ac{Thank you for catching that} 
% From the examples, we can see the input KG components (in bold) are verbalized in the explanations, where each explanation is fluent, specific, and objective to the current item. In the first example, which is about the novel \enquote{Flashback}, the explanation contains author and award information, where the generated output \textit{hallucinates} extra information about where the author is from~\cite{rohrbach-etal-2018-object}. The second example, which is from Movie KG-Exp about the movie \enquote{Journey to the Center of the Earth} explains the recommendation through item information about the original source material of the movie (\enquote{classic novel}). Here, the model adds \enquote{center of the earth} which may come from the movie's title itself, but misses some minute details presented in the ground truth. However, both demonstrate our model’s capability in incorporating features from the user-item KG.
\section{Conclusion}\label{sec:conc}
We propose KnowRec, a Knowledge-aware model for generating NL explanations and recommendation scores on user-item pairs. 
% KnowRec adapts a collaborative knowledge representation approach adapted on state-of-the-art KG-to-text models. 
To evaluate KnowRec, we devise and release a semi-supervised large-scale KG-NL recommendation dataset in the book and movie domain. Extensive experiments on both datasets demonstrate the suitability of our model compared to recently proposed explainable recommendation models. We hope that by proposing this KG-guided task, we will open up avenues to research focused on detailed, objective, and specific explanations which can also scale to new items and users, rather than the current review-focused work. In future work, we plan to incorporate user-specific KGs and other pre-trained language models into our model in order to verbalize both user and item-level feature explanations.  
\section{Limitations}\label{sec:limit}
%Mention we cannot yet handle zero-purchase (user without purchase history) problem?  given our data
%Mention we only have computational resources for one LM (future work)
%Mention limitations of the dataset
%book vs movie description. for example the book kgs are typically author related and thus capture more information than the movie data
While our approach generates objective, descriptive explanations while implicitly capturing personalized aspects of a user's purchase history, currently our dataset labels are limited to item-specific explanations, with the book-related KGs typically containing author-related information, and thus more information dense than the movie-related KGs. 
% Second, our datasets are extracted from publically available KGs which are often incomplete. 
These limitations are due to the currently available datasets, and future work can explore constructing a more personalized user-item KG for explainable recommendation. Furthermore, in our approach, we represent users through their item purchase history. Therefore, while we handle the zero-purchase case for items (items that have not been purchased before), the zero-purchase case for users (users without a purchase history) is outside the scope of our work. In the future, we will extend our approach to user-attributed datasets to handle such cases.
\section{Ethics Statement}\label{sec:ethics}
All our experiments are performed over publicly available datasets. We do not use any identifiable information about crowd workers who provide annotations for these datasets. Neither do we perform any additional annotations or human evaluations in this work. We do not foresee any risks using KnowRec if the inputs to our model are designed as per our procedure. However, our models may exhibit unwanted biases that are inherent in pre-trained language models. This aspect is beyond the scope of the current work.

\bibliography{anthology,custom}

\begin{thebibliography}{50}
\expandafter\ifx\csname natexlab\endcsname\relax\def\natexlab#1{#1}\fi

\bibitem[{Asghar(2016)}]{asghar2016yelp}
Nabiha Asghar. 2016.
\newblock \href {https://arxiv.org/abs/1605.05362} {Yelp dataset challenge:
  Review rating prediction}.
\newblock \emph{arXiv preprint arXiv:1605.05362}.

\bibitem[{Auer et~al.(2007)Auer, Bizer, Kobilarov, Lehmann, Cyganiak, and
  Ives}]{auer2007dbpedia}
S\"{o}ren Auer, Christian Bizer, Georgi Kobilarov, Jens Lehmann, Richard
  Cyganiak, and Zachary Ives. 2007.
\newblock \href {https://dl.acm.org/doi/10.5555/1785162.1785216} {{DBpedia}:
  {A} nucleus for a web of open data}.
\newblock In \emph{Proceedings of the 6th International The Semantic Web and
  2nd Asian Conference on Asian Semantic Web Conference}, pages 722--735,
  Berlin, Heidelberg. Springer-Verlag.

\bibitem[{Chen et~al.(2018)Chen, Zhang, Liu, and Ma}]{chen2018neural}
Chong Chen, Min Zhang, Yiqun Liu, and Shaoping Ma. 2018.
\newblock \href {https://dl.acm.org/doi/10.1145/3178876.3186070} {Neural
  attentional rating regression with review-level explanations}.
\newblock In \emph{Proceedings of the 2018 World Wide Web Conference}, pages
  1583--1592.

\bibitem[{Chen et~al.(2021)Chen, Shi, Li, and Zhang}]{chen2021neural}
Hanxiong Chen, Shaoyun Shi, Yunqi Li, and Yongfeng Zhang. 2021.
\newblock Neural collaborative reasoning.
\newblock In \emph{Proceedings of the Web Conference 2021}, pages 1516--1527.

\bibitem[{Chen et~al.(2020)Chen, Su, Yan, and Wang}]{chen-etal-2020-kgpt}
Wenhu Chen, Yu~Su, Xifeng Yan, and William~Yang Wang. 2020.
\newblock \href {https://doi.org/10.18653/v1/2020.emnlp-main.697} {{KGPT}:
  Knowledge-grounded pre-training for data-to-text generation}.
\newblock In \emph{Proceedings of the 2020 Conference on Empirical Methods in
  Natural Language Processing (EMNLP)}, pages 8635--8648, Online. Association
  for Computational Linguistics.

\bibitem[{Colas et~al.(2022)Colas, Alvandipour, and Wang}]{colas-etal-2022-gap}
Anthony Colas, Mehrdad Alvandipour, and Daisy~Zhe Wang. 2022.
\newblock \href {https://aclanthology.org/2022.coling-1.506} {{GAP}: A
  graph-aware language model framework for knowledge graph-to-text generation}.
\newblock In \emph{Proceedings of the 29th International Conference on
  Computational Linguistics}, pages 5755--5769, Gyeongju, Republic of Korea.
  International Committee on Computational Linguistics.

\bibitem[{Dong et~al.(2017)Dong, Huang, Wei, Lapata, Zhou, and
  Xu}]{dong-etal-2017-learning-generate}
Li~Dong, Shaohan Huang, Furu Wei, Mirella Lapata, Ming Zhou, and Ke~Xu. 2017.
\newblock \href {https://aclanthology.org/E17-1059} {Learning to generate
  product reviews from attributes}.
\newblock In \emph{Proceedings of the 15th Conference of the {E}uropean Chapter
  of the Association for Computational Linguistics: Volume 1, Long Papers},
  pages 623--632, Valencia, Spain. Association for Computational Linguistics.

\bibitem[{Du et~al.(2022)Du, Zhu, Chen, Zheng, and Gao}]{du2022hakg}
Yuntao Du, Xinjun Zhu, Lu~Chen, Baihua Zheng, and Yunjun Gao. 2022.
\newblock \href {https://arxiv.org/abs/2204.04959} {{HAKG}: {H}ierarchy-aware
  knowledge gated network for recommendation}.
\newblock \emph{arXiv preprint arXiv:2204.04959}.

\bibitem[{Fu et~al.(2020)Fu, Xian, Gao, Zhao, Huang, Ge, Xu, Geng, Shah, Zhang
  et~al.}]{fu2020fairness}
Zuohui Fu, Yikun Xian, Ruoyuan Gao, Jieyu Zhao, Qiaoying Huang, Yingqiang Ge,
  Shuyuan Xu, Shijie Geng, Chirag Shah, Yongfeng Zhang, et~al. 2020.
\newblock Fairness-aware explainable recommendation over knowledge graphs.
\newblock In \emph{Proceedings of the 43rd International ACM SIGIR Conference
  on Research and Development in Information Retrieval}, pages 69--78.

\bibitem[{Gardent et~al.(2017)Gardent, Shimorina, Narayan, and
  Perez-Beltrachini}]{gardent-etal-2017-webnlg}
Claire Gardent, Anastasia Shimorina, Shashi Narayan, and Laura
  Perez-Beltrachini. 2017.
\newblock \href {https://doi.org/10.18653/v1/W17-3518} {The {W}eb{NLG}
  challenge: Generating text from {RDF} data}.
\newblock In \emph{Proceedings of the 10th International Conference on Natural
  Language Generation}, pages 124--133, Santiago de Compostela, Spain.
  Association for Computational Linguistics.

\bibitem[{Hamilton et~al.(2017)Hamilton, Ying, and
  Leskovec}]{hamilton2017inductive}
Will Hamilton, Zhitao Ying, and Jure Leskovec. 2017.
\newblock \href
  {https://proceedings.neurips.cc/paper/2017/file/5dd9db5e033da9c6fb5ba83c7a7ebea9-Paper.pdf}
  {Inductive representation learning on large graphs}.
\newblock In \emph{Advances in Neural Information Processing Systems},
  volume~30.

\bibitem[{He and McAuley(2016)}]{he2016ups}
Ruining He and Julian McAuley. 2016.
\newblock \href {https://dl.acm.org/doi/10.1145/2872427.2883037} {Ups and
  downs: Modeling the visual evolution of fashion trends with one-class
  collaborative filtering}.
\newblock In \emph{Proceedings of the 25th international conference on World
  Wide Web}, pages 507--517.

\bibitem[{Hokamp and Liu(2017)}]{hokamp-liu-2017-lexically}
Chris Hokamp and Qun Liu. 2017.
\newblock \href {https://doi.org/10.18653/v1/P17-1141} {Lexically constrained
  decoding for sequence generation using grid beam search}.
\newblock In \emph{Proceedings of the 55th Annual Meeting of the Association
  for Computational Linguistics (Volume 1: Long Papers)}, pages 1535--1546,
  Vancouver, Canada. Association for Computational Linguistics.

\bibitem[{Hou et~al.(2019)Hou, Wu, Chen, Li, Zheng, and
  Liu}]{hou2019explainable}
Min Hou, Le~Wu, Enhong Chen, Zhi Li, Vincent~W Zheng, and Qi~Liu. 2019.
\newblock \href {https://dl.acm.org/doi/10.5555/3367471.3367694} {Explainable
  fashion recommendation: {A} semantic attribute region guided approach}.
\newblock In \emph{Proceedings of the 28th International Joint Conference on
  Artificial Intelligence}, pages 4681--4688.

\bibitem[{Hui et~al.(2022)Hui, Zhang, Zhou, Wen, and
  Nian}]{hui2022personalized}
Bei Hui, Lizong Zhang, Xue Zhou, Xiao Wen, and Yuhui Nian. 2022.
\newblock \href {https://link.springer.com/article/10.1007/s10489-021-02363-w}
  {Personalized recommendation system based on knowledge embedding and
  historical behavior}.
\newblock \emph{Applied Intelligence}, 52(1):954--966.

\bibitem[{Ke et~al.(2021)Ke, Ji, Ran, Cui, Wang, Song, Zhu, and
  Huang}]{ke-etal-2021-jointgt}
Pei Ke, Haozhe Ji, Yu~Ran, Xin Cui, Liwei Wang, Linfeng Song, Xiaoyan Zhu, and
  Minlie Huang. 2021.
\newblock \href {https://doi.org/10.18653/v1/2021.findings-acl.223}
  {{J}oint{GT}: Graph-text joint representation learning for text generation
  from knowledge graphs}.
\newblock In \emph{Findings of the Association for Computational Linguistics:
  ACL-IJCNLP 2021}, pages 2526--2538, Online. Association for Computational
  Linguistics.

\bibitem[{Kingma and Ba(2015)}]{kingma2015adam}
Diederik~P Kingma and Jimmy Ba. 2015.
\newblock \href {https://openreview.net/forum?id=8gmWwjFyLj} {Adam: A method
  for stochastic optimization}.
\newblock In \emph{Proceedings of the International Conference on Learning
  Representations}.

\bibitem[{Koncel-Kedziorski et~al.(2019)Koncel-Kedziorski, Bekal, Luan, Lapata,
  and Hajishirzi}]{koncel-kedziorski-etal-2019-text}
Rik Koncel-Kedziorski, Dhanush Bekal, Yi~Luan, Mirella Lapata, and Hannaneh
  Hajishirzi. 2019.
\newblock \href {https://doi.org/10.18653/v1/N19-1238} {{T}ext {G}eneration
  from {K}nowledge {G}raphs with {G}raph {T}ransformers}.
\newblock In \emph{Proceedings of the 2019 Conference of the North {A}merican
  Chapter of the Association for Computational Linguistics: Human Language
  Technologies, Volume 1 (Long and Short Papers)}, pages 2284--2293,
  Minneapolis, Minnesota. Association for Computational Linguistics.

\bibitem[{Koren(2008)}]{koren2008factorization}
Yehuda Koren. 2008.
\newblock \href {https://dl.acm.org/doi/abs/10.1145/1401890.1401944}
  {Factorization meets the neighborhood: a multifaceted collaborative filtering
  model}.
\newblock In \emph{Proceedings of the 14th ACM SIGKDD international conference
  on Knowledge discovery and data mining}, pages 426--434.

\bibitem[{Lewis et~al.(2020)Lewis, Liu, Goyal, Ghazvininejad, Mohamed, Levy,
  Stoyanov, and Zettlemoyer}]{lewis-etal-2020-bart}
Mike Lewis, Yinhan Liu, Naman Goyal, Marjan Ghazvininejad, Abdelrahman Mohamed,
  Omer Levy, Veselin Stoyanov, and Luke Zettlemoyer. 2020.
\newblock \href {https://doi.org/10.18653/v1/2020.acl-main.703} {{BART}:
  Denoising sequence-to-sequence pre-training for natural language generation,
  translation, and comprehension}.
\newblock In \emph{Proceedings of the 58th Annual Meeting of the Association
  for Computational Linguistics}, pages 7871--7880, Online. Association for
  Computational Linguistics.

\bibitem[{Li et~al.(2019)Li, Quan, Peng, Qi, Deng, and Wu}]{li2019capsule}
Chenliang Li, Cong Quan, Li~Peng, Yunwei Qi, Yuming Deng, and Libing Wu. 2019.
\newblock \href {https://dl.acm.org/doi/10.1145/3331184.3331216} {A capsule
  network for recommendation and explaining what you like and dislike}.
\newblock In \emph{Proceedings of the 42nd International ACM SIGIR conference
  on Research and Development in Information Retrieval}, pages 275--284.

\bibitem[{Li et~al.(2020)Li, Zhang, and Chen}]{li2020generate}
Lei Li, Yongfeng Zhang, and Li~Chen. 2020.
\newblock \href {https://dl.acm.org/doi/10.1145/3340531.3411992} {Generate
  neural template explanations for recommendation}.
\newblock In \emph{Proceedings of the 29th ACM International Conference on
  Information \& Knowledge Management}, pages 755--764.

\bibitem[{Li et~al.(2021)Li, Zhang, and Chen}]{li-etal-2021-personalized}
Lei Li, Yongfeng Zhang, and Li~Chen. 2021.
\newblock \href {https://doi.org/10.18653/v1/2021.acl-long.383} {Personalized
  transformer for explainable recommendation}.
\newblock In \emph{Proceedings of the 59th Annual Meeting of the Association
  for Computational Linguistics and the 11th International Joint Conference on
  Natural Language Processing (Volume 1: Long Papers)}, pages 4947--4957,
  Online. Association for Computational Linguistics.

\bibitem[{Li et~al.(2022)Li, Zhang, and Chen}]{li2022personalized}
Lei Li, Yongfeng Zhang, and Li~Chen. 2022.
\newblock \href {https://arxiv.org/abs/2202.07371} {Personalized prompt
  learning for explainable recommendation}.
\newblock \emph{arXiv preprint arXiv:2202.07371}.

\bibitem[{Li et~al.(2017)Li, Wang, Ren, Bing, and Lam}]{li2017neural}
Piji Li, Zihao Wang, Zhaochun Ren, Lidong Bing, and Wai Lam. 2017.
\newblock \href {https://dl.acm.org/doi/10.1145/3077136.3080822} {Neural rating
  regression with abstractive tips generation for recommendation}.
\newblock In \emph{Proceedings of the 40th International ACM SIGIR conference
  on Research and Development in Information Retrieval}, pages 345--354.

\bibitem[{Lin(2004)}]{lin-2004-rouge}
Chin-Yew Lin. 2004.
\newblock \href {https://aclanthology.org/W04-1013} {{ROUGE}: A package for
  automatic evaluation of summaries}.
\newblock In \emph{Text Summarization Branches Out}, pages 74--81, Barcelona,
  Spain. Association for Computational Linguistics.

\bibitem[{Ma et~al.(2019)Ma, Zhang, Cao, Jin, Wang, Liu, Ma, and
  Ren}]{ma2019jointly}
Weizhi Ma, Min Zhang, Yue Cao, Woojeong Jin, Chenyang Wang, Yiqun Liu, Shaoping
  Ma, and Xiang Ren. 2019.
\newblock Jointly learning explainable rules for recommendation with knowledge
  graph.
\newblock In \emph{The world wide web conference}, pages 1210--1221.

\bibitem[{Mendes et~al.(2011)Mendes, Jakob, Garc{\'\i}a-Silva, and
  Bizer}]{mendes2011dbpedia}
Pablo~N Mendes, Max Jakob, Andr{\'e}s Garc{\'\i}a-Silva, and Christian Bizer.
  2011.
\newblock \href {https://dl.acm.org/doi/10.1145/2063518.2063519} {{DBpedia}
  spotlight: {S}hedding light on the web of documents}.
\newblock In \emph{Proceedings of the 7th international conference on semantic
  systems}, pages 1--8.

\bibitem[{Mnih and Salakhutdinov(2007)}]{mnih2007probabilistic}
Andriy Mnih and Russ~R Salakhutdinov. 2007.
\newblock \href
  {https://proceedings.neurips.cc/paper/2007/file/d7322ed717dedf1eb4e6e52a37ea7bcd-Paper.pdf}
  {Probabilistic matrix factorization}.
\newblock In \emph{Advances in Neural Information Processing Systems},
  volume~20.

\bibitem[{Papineni et~al.(2002)Papineni, Roukos, Ward, and
  Zhu}]{papineni-etal-2002-bleu}
Kishore Papineni, Salim Roukos, Todd Ward, and Wei-Jing Zhu. 2002.
\newblock \href {https://doi.org/10.3115/1073083.1073135} {{B}leu: a method for
  automatic evaluation of machine translation}.
\newblock In \emph{Proceedings of the 40th Annual Meeting of the Association
  for Computational Linguistics}, pages 311--318, Philadelphia, Pennsylvania,
  USA. Association for Computational Linguistics.

\bibitem[{Radford et~al.(2019)Radford, Wu, Child, Luan, Amodei, Sutskever
  et~al.}]{radford2019language}
Alec Radford, Jeffrey Wu, Rewon Child, David Luan, Dario Amodei, Ilya
  Sutskever, et~al. 2019.
\newblock \href {https://openai.com/blog/better-language-models/} {Language
  models are unsupervised multitask learners}.
\newblock \emph{OpenAI blog}, 1(8).

\bibitem[{Ribeiro et~al.(2021)Ribeiro, Schmitt, Sch{\"u}tze, and
  Gurevych}]{ribeiro-etal-2021-investigating}
Leonardo F.~R. Ribeiro, Martin Schmitt, Hinrich Sch{\"u}tze, and Iryna
  Gurevych. 2021.
\newblock \href {https://doi.org/10.18653/v1/2021.nlp4convai-1.20}
  {Investigating pretrained language models for graph-to-text generation}.
\newblock In \emph{Proceedings of the 3rd Workshop on Natural Language
  Processing for Conversational AI}, pages 211--227, Online. Association for
  Computational Linguistics.

\bibitem[{Shi et~al.(2020)Shi, Chen, Ma, Mao, Zhang, and Zhang}]{shi2020neural}
Shaoyun Shi, Hanxiong Chen, Weizhi Ma, Jiaxin Mao, Min Zhang, and Yongfeng
  Zhang. 2020.
\newblock Neural logic reasoning.
\newblock In \emph{Proceedings of the 29th ACM International Conference on
  Information \& Knowledge Management}, pages 1365--1374.

\bibitem[{Sun et~al.(2020)Sun, Wu, Zhang, Fu, Hong, and
  Wang}]{10.1145/3366423.3380164}
Peijie Sun, Le~Wu, Kun Zhang, Yanjie Fu, Richang Hong, and Meng Wang. 2020.
\newblock \href {https://doi.org/10.1145/3366423.3380164} {Dual learning for
  explainable recommendation: Towards unifying user preference prediction and
  review generation}.
\newblock In \emph{Proceedings of The Web Conference 2020}, WWW '20, page
  837–847, New York, NY, USA. Association for Computing Machinery.

\bibitem[{Tintarev and Masthoff(2015)}]{tintarev2015explaining}
Nava Tintarev and Judith Masthoff. 2015.
\newblock \href
  {https://link.springer.com/chapter/10.1007/978-1-4899-7637-6_10} {Explaining
  recommendations: Design and evaluation}.
\newblock In \emph{Recommender systems handbook}, pages 353--382. Springer.

\bibitem[{Vaswani et~al.(2017)Vaswani, Shazeer, Parmar, Uszkoreit, Jones,
  Gomez, Kaiser, and Polosukhin}]{vaswani2017attention}
Ashish Vaswani, Noam Shazeer, Niki Parmar, Jakob Uszkoreit, Llion Jones,
  Aidan~N Gomez, \L~ukasz Kaiser, and Illia Polosukhin. 2017.
\newblock \href
  {https://proceedings.neurips.cc/paper/2017/file/3f5ee243547dee91fbd053c1c4a845aa-Paper.pdf}
  {Attention is all you need}.
\newblock In \emph{Advances in Neural Information Processing Systems},
  volume~30.

\bibitem[{Veli{\v{c}}kovi{\'c} et~al.(2018)Veli{\v{c}}kovi{\'c}, Cucurull,
  Casanova, Romero, Li{\`o}, and Bengio}]{velivckovic2018graph}
Petar Veli{\v{c}}kovi{\'c}, Guillem Cucurull, Arantxa Casanova, Adriana Romero,
  Pietro Li{\`o}, and Yoshua Bengio. 2018.
\newblock \href {https://openreview.net/forum?id=rJXMpikCZ} {Graph attention
  networks}.
\newblock In \emph{Proceedings of the International Conference on Learning
  Representations}.

\bibitem[{Wang et~al.(2018{\natexlab{a}})Wang, Wang, Jia, and
  Yin}]{wang2018explainable}
Nan Wang, Hongning Wang, Yiling Jia, and Yue Yin. 2018{\natexlab{a}}.
\newblock \href {https://dl.acm.org/doi/abs/10.1145/3209978.3210010}
  {Explainable recommendation via multi-task learning in opinionated text
  data}.
\newblock In \emph{The 41st International ACM SIGIR Conference on Research \&
  Development in Information Retrieval}, pages 165--174.

\bibitem[{Wang et~al.(2019)Wang, He, Cao, Liu, and Chua}]{wang2019kgat}
Xiang Wang, Xiangnan He, Yixin Cao, Meng Liu, and Tat-Seng Chua. 2019.
\newblock \href {https://dl.acm.org/doi/10.1145/3292500.3330989} {{KGAT}:
  {K}nowledge graph attention network for recommendation}.
\newblock In \emph{Proceedings of the 25th ACM SIGKDD international conference
  on knowledge discovery \& data mining}, pages 950--958.

\bibitem[{Wang et~al.(2021)Wang, Huang, Wang, Yuan, Liu, He, and
  Chua}]{wang2021learning}
Xiang Wang, Tinglin Huang, Dingxian Wang, Yancheng Yuan, Zhenguang Liu,
  Xiangnan He, and Tat-Seng Chua. 2021.
\newblock \href {https://dl.acm.org/doi/10.1145/3442381.3450133} {Learning
  intents behind interactions with knowledge graph for recommendation}.
\newblock In \emph{Proceedings of the Web Conference 2021}, pages 878--887.

\bibitem[{Wang et~al.(2018{\natexlab{b}})Wang, Chen, Yang, Wu, Wu, and
  Xie}]{wang2018reinforcement}
Xiting Wang, Yiru Chen, Jie Yang, Le~Wu, Zhengtao Wu, and Xing Xie.
  2018{\natexlab{b}}.
\newblock \href {https://ieeexplore.ieee.org/document/8594883} {A reinforcement
  learning framework for explainable recommendation}.
\newblock In \emph{2018 IEEE International Conference on Data Mining}, pages
  587--596. IEEE.

\bibitem[{Wang et~al.(2020)Wang, Lin, Tan, Chen, and Liu}]{wang2020ckan}
Ze~Wang, Guangyan Lin, Huobin Tan, Qinghong Chen, and Xiyang Liu. 2020.
\newblock \href {https://dl.acm.org/doi/10.1145/3397271.3401141} {{CKAN}:
  {C}ollaborative knowledge-aware attentive network for recommender systems}.
\newblock In \emph{Proceedings of the 43rd International ACM SIGIR conference
  on Research and Development in Information Retrieval}, pages 219--228.

\bibitem[{Welling and Kipf(2016)}]{welling2016semi}
Max Welling and Thomas~N Kipf. 2016.
\newblock \href {https://openreview.net/forum?id=SJU4ayYgl} {Semi-supervised
  classification with graph convolutional networks}.
\newblock In \emph{Proceedings of the International Conference on Learning
  Representations}.

\bibitem[{Wu et~al.(2018)Wu, Dai, Yin, Huang, and Chen}]{wu2018improving}
Zhen Wu, Xin-Yu Dai, Cunyan Yin, Shujian Huang, and Jiajun Chen. 2018.
\newblock \href {https://ojs.aaai.org/index.php/AAAI/article/view/12054}
  {Improving review representations with user attention and product attention
  for sentiment classification}.
\newblock \emph{Proceedings of the AAAI Conference on Artificial Intelligence},
  32(1).

\bibitem[{Xian et~al.(2019)Xian, Fu, Muthukrishnan, De~Melo, and
  Zhang}]{xian2019reinforcement}
Yikun Xian, Zuohui Fu, Shan Muthukrishnan, Gerard De~Melo, and Yongfeng Zhang.
  2019.
\newblock Reinforcement knowledge graph reasoning for explainable
  recommendation.
\newblock In \emph{Proceedings of the 42nd international ACM SIGIR conference
  on research and development in information retrieval}, pages 285--294.

\bibitem[{Xie et~al.(2021)Xie, Hu, Cai, Zhang, and Chen}]{xie2021explainable}
Lijie Xie, Zhaoming Hu, Xingjuan Cai, Wensheng Zhang, and Jinjun Chen. 2021.
\newblock Explainable recommendation based on knowledge graph and
  multi-objective optimization.
\newblock \emph{Complex \& Intelligent Systems}, 7(3):1241--1252.

\bibitem[{Yang et~al.(2021)Yang, Wang, Deng, and Wang}]{yang2021explanation}
Aobo Yang, Nan Wang, Hongbo Deng, and Hongning Wang. 2021.
\newblock \href {https://dl.acm.org/doi/abs/10.1145/3437963.3441726}
  {Explanation as a defense of recommendation}.
\newblock In \emph{Proceedings of the 14th ACM International Conference on Web
  Search and Data Mining}, pages 1029--1037.

\bibitem[{Yu et~al.(2022)Yu, Zhu, Li, Hu, Wang, Ji, and Jiang}]{yu2022survey}
Wenhao Yu, Chenguang Zhu, Zaitang Li, Zhiting Hu, Qingyun Wang, Heng Ji, and
  Meng Jiang. 2022.
\newblock A survey of knowledge-enhanced text generation.
\newblock \emph{ACM Computing Surveys}, 54(11s):1--38.

\bibitem[{Zhang et~al.(2020)Zhang, Chen et~al.}]{zhang2020explainable}
Yongfeng Zhang, Xu~Chen, et~al. 2020.
\newblock \href {https://dl.acm.org/doi/10.1561/1500000066} {Explainable
  recommendation: A survey and new perspectives}.
\newblock \emph{Foundations and Trends{\textregistered} in Information
  Retrieval}, 14(1):1--101.

\bibitem[{Zhu et~al.(2021)Zhu, Xian, Fu, de~Melo, and
  Zhang}]{zhu-etal-2021-faithfully}
Yaxin Zhu, Yikun Xian, Zuohui Fu, Gerard de~Melo, and Yongfeng Zhang. 2021.
\newblock \href {https://doi.org/10.18653/v1/2021.naacl-main.245} {Faithfully
  explainable recommendation via neural logic reasoning}.
\newblock In \emph{Proceedings of the 2021 Conference of the North American
  Chapter of the Association for Computational Linguistics: Human Language
  Technologies}, pages 3083--3090, Online. Association for Computational
  Linguistics.

\end{thebibliography}
\bibliographystyle{acl_natbib}

\appendix

\section{Dataset Details}
\subsection{Source Data}
\textbf{Amazon product data:} The Amazon product dataset is a large-scale widely used dataset for product recommendation containing product reviews and metadata from Amazon. Data fields include ratings, texts, descriptions, and category information~\cite{he2016ups}. Because the dataset contains item descriptions, we can leverage such data to extract entities and relations to construct a KG that matches the textual description. Thus, these descriptions provide objective, item-distinct explanations as to why a user may have purchased a product. Although a user may not have reviewed an item, the dataset provides an existing description of the item, allowing models to produce explanations for such items. To keep our datasets large-scale, we focus on Amazon Book and Amazon Movie 5-core, the two largest Amazon product datasets. 

\subsection{Dataset Comparison}
\label{sec:dataset-comparison}
\begin{table*}[!ht]
\centering
\small
% \resizebox{\textwidth}{!}{%
\setlength\tabcolsep{5pt}
\begin{tabular}{%r|
lrrrlrrrlr
%l|lllllllll
}
\toprule
%\hline
%Task &
  \multicolumn{1}{c}{Name} &
  \multicolumn{1}{c}{\#Users} &
  \multicolumn{1}{c}{\#Items} &
  \multicolumn{1}{c}{\#Interactions} &
  \multicolumn{1}{c}{KG} &
  \multicolumn{1}{c}{\#Es} &
  \multicolumn{1}{c}{\#Rs} &
  \multicolumn{1}{c}{\#Triples} &
  \multicolumn{1}{c}{Desc.} &
  \begin{tabular}[c]{@{}l@{}}Words/\\Sample \end{tabular} \\
\midrule
%\hline
%KG-Rec  & 
Last.FM             & 23,566  & 48,123  & 3,034,796  & Yes & 58,266  & 9  & 464,567   & No & -    \\
%KG-Rec  & 
Book-Crossing       & 276,271 & 271,379 & 1,048,575  & Yes & 25,787  & 18 & 60,787    & No & -    \\
%KG-Rec  & 
Movie-Lens20M       & 138,159 & 16,954  & 13,501,622 & Yes & 102,569 & 32 & 499,474   & No & -    \\
%KG-Rec  & 
Amazon-book (KG)    & 70,679  & 24,915  & 847,733    & Yes & 88,572  & 39 & 2,557,746 & No & -    \\ %\midrule
\hline
%Exp-Rec & 
Yelp-Restaurant     & 27,147  & 20,266  & 1,293,247  & No  & -      & - & -        & No & 12.32 \\
%Exp-Rec & 
Amazon Movies & 7,506   & 7,360   & 441,783    & No  & -      & - & -        & No & 14.14 \\
%Exp-Rec & 
TripAdvisor-Hotel   & 9,765   & 6,280   & 320,023    & No  & -      & - & -        & No & 13.01 \\ 
%\midrule
\hline
%\textit{KG-Exp} &
  \textit{Book KG-Exp} &
  \textit{396,114} &
  \textit{95,733} &
  \textit{2,318,107} &
  \textit{Yes} &
  \textit{195,110} &
  \textit{392} &
  \textit{745,699} &
  \textit{Yes} &
  \textit{99.96} \\
%\textit{KG-Exp} &
  \textit{Movie KG-Exp} &
  \textit{131,375} &
  \textit{18,107} &
  \textit{788,957} &
  \textit{Yes} &
  \textit{59,036} &
  \textit{363} &
  \textit{146,772} &
  \textit{Yes} &
  \textit{96.35} \\
\bottomrule
%\hline
\end{tabular}%
% }
\caption{\label{tab:datacompare}Comparison of widely used datasets divided by task: KG-Recommendation (top), Explainable Recommendation (middle), and KG Explainable Recommendation (bottom).%\jun{Please explicitly explain which of KG-REC, Exp-Rec, and KG-Exp corresponds to which of the three categories separated by horizontal lines.}\ac{added context to the caption above}\jun{Looks good!}
% \textit{\#Es}, \textit{\#Rs}, and \textit{Desc.} denote number of entities, number of relations, and if the datset contains parallel descriptions.
%\jun{I think a dataset's name is good if it is as short as possible and clear about its origin (if any). Given that, I wonder if \enquote{AMZ-Book KG-Exp} or more simply \enquote{Book KG-Exp} is slightly better. What do you think?}\ac{I like Book KG-Exp and will change it. Thank you.}
}
% \vspace{-0.4cm}
\end{table*}
Table~\ref{tab:datacompare} summarizes existing popular recommendation system datasets utilized for both the explainable recommendation and KG recommendation task. We report both traditional recommendation features, KG-recommendation features, and explainable recommendation features. Last.FM~\cite{wang2019kgat}, Book-Crossing~\cite{wang2020ckan}, Movie-Lens20M~\cite{wang2020ckan}, and Amazon-book (KG)~\cite{wang2019kgat} are popular benchmarks for the KG-recommendation task but contain no NL explanation features. Yelp-Restaurant, Amazon Movies \& TV, and TripAdvisor-Hotel have been recently experimented with for the explainable recommendation task~\cite{li2020generate}, but lack KG data and rely on user reviews as proxies for the explanation. In contrast, our datasets, referred to as \textit{Book KG-Exp} and \textit{Movie KG-Exp} contain both KG and the corresponding parallel item descriptions associated with those KGs as explanations. Compared to Book KG-Exp, the Movie KG-Exp dataset contains fewer amount of unique KG elements, with \textit{59,036} to \textit{195,110} and \textit{745,699} to \textit{146,772} unique entities and KG, while having similarly sized explanations. 

\subsection{Dataset Statistics}
\begin{figure*}[h]
  \centering
  
  \subfigure[Book KGs]{\label{fig:bookkg}\includegraphics[width=0.24\textwidth]{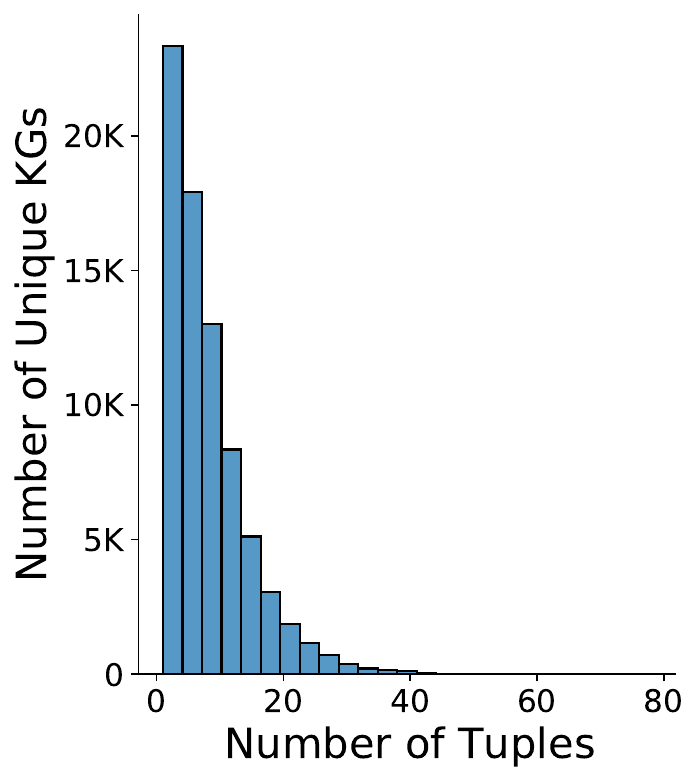}} 
  \subfigure[Movie KGs]{\label{fig:moviekg}\includegraphics[width=0.24\textwidth]{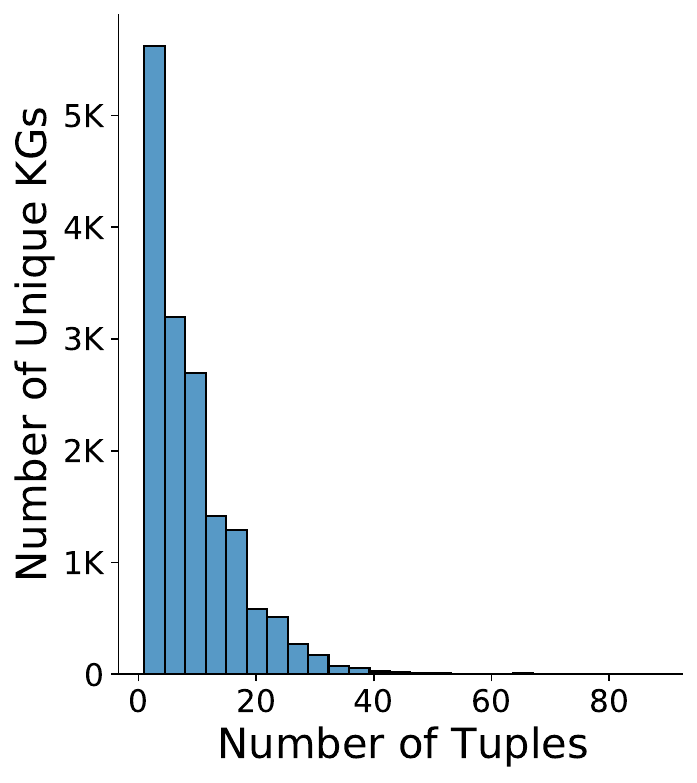}} 
  \subfigure[Book Explanations]{\label{fig:bookexplain}\includegraphics[width=0.24\textwidth]{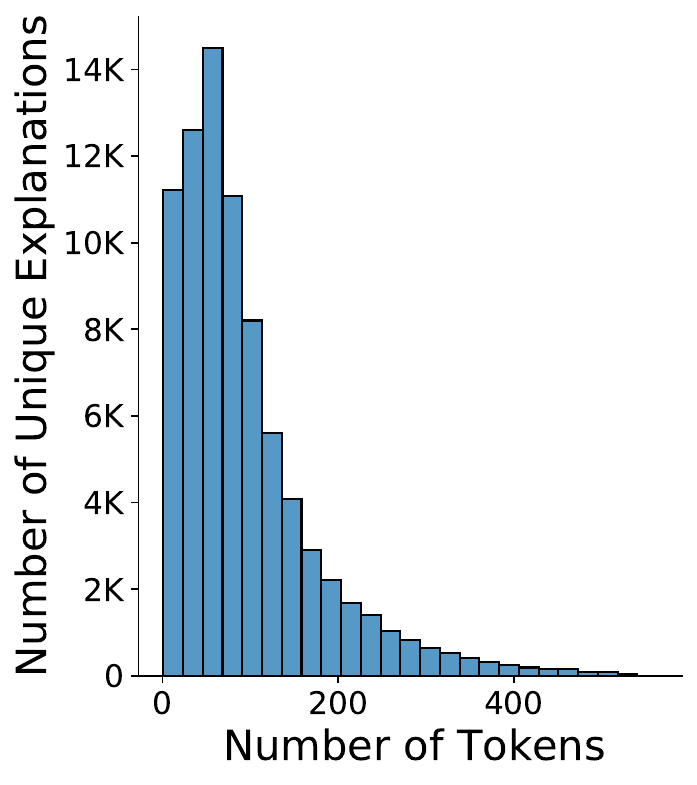}}
  \subfigure[Movie Explanations]{\label{fig:movieexplain}\includegraphics[width=0.24\textwidth]{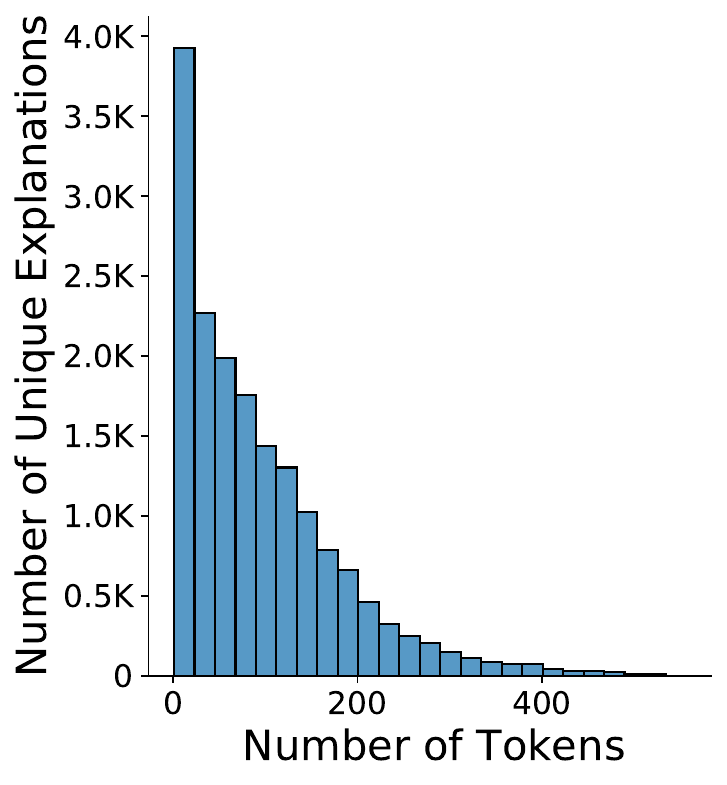}}

  \caption{Distributions for number of tuples (\cref{fig:bookkg,fig:moviekg}) and tokens (\cref{fig:bookexplain,fig:movieexplain}) per sample.}
 
\label{fig:stats}
\end{figure*}

We provide detailed statistics on both the Book KG-Exp and Movie KG-Exp datasets in \cref{fig:stats}.
As seen in \cref{fig:bookkg,fig:moviekg}, the distributions of KGs with respect to the number of tuples %, where both contain
shows similar long-tail distributions in both datasets. We observe from \cref{fig:bookexplain,fig:movieexplain} that a similar trend of long-tail distributions exists for both with respect to explanation lengths, where the lengths in the book dataset tend to skew more right than the lengths in the movie dataset.

\section{Experiment Details}
\label{sec:settings}
\subsection{Baseline Models}
\label{sec:baselines}
We introduce several baselines in explainable recommendation, describing how to adapt the models to the KG setting, as these models have been primarily formulated for user review data.

\textbf{Att2Seq}~\cite{dong-etal-2017-learning-generate} was designed for review generation, where we adapt it to the item explanation setting. As in~\cite{li-etal-2021-personalized}, we remove the attention module, as it makes the generated content unreadable.%\jun{I think introducing a baseline's name in bold font is a good idea, but we should keep the same indentation as other paragraphs.}\ac{No problem, I have now indented them.}

\textbf{NRT}~\cite{li2017neural} is a multi-task model for rating prediction and tip generation, based on user and item IDs. As in previous work, we use our explanations as tips and remove the model's L2 regularizer~\cite{li2020generate, li-etal-2021-personalized}, which causes the model to generate identical sentences.

\textbf{Transformer}~\cite{vaswani2017attention,li-etal-2021-personalized} treats user and item IDs as words. We adapt the model first introduced for review generation by~\citet{li-etal-2021-personalized} while integrating the KG entities and relations instead of the review item features.

\textbf{PETER}~\cite{li-etal-2021-personalized} utilizes both user/item IDs and corresponding item features extracted from user reviews to generate a recommendation score, explanation, and context related to the item features. The model also develops a novel PETER mask between item/user IDs and corresponding features/generated text. As our task does not take a feature-based approach, for a fair comparison we remove the context prediction module and input the whole KG into the model as the corresponding item features.

\textbf{PEPLER}~\cite{li2022personalized} is an extension of PETER, where the transformer is replaced with a pre-train language model, namely GPT-2 to generate both recommendation scores and explanations. We take the best-performing setting for a fair comparison, namely using the MLP setting for recommendation scores. 

In addition to NRT, PETER, and PEPLER, as in previous work, we compare with two traditional baselines for recommendation: \textbf{PMF}~\cite{mnih2007probabilistic} and \textbf{SVD++}~\cite{koren2008factorization}.

\subsection{Hyper-parameters and Settings}
\label{sec:hyper-settings}
\begin{figure}[h]
\centering
\includegraphics[width=\columnwidth]{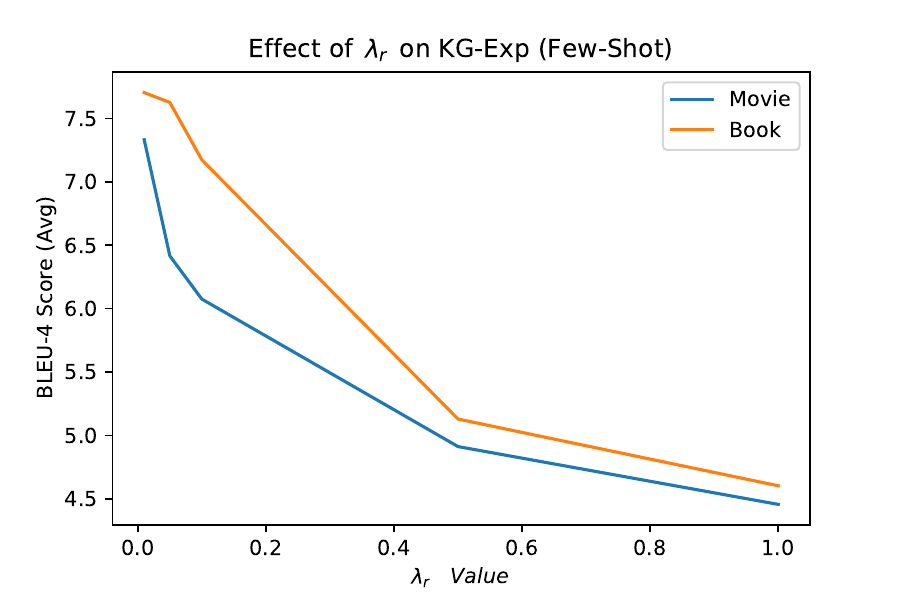}
\caption{Effect of $\lambda_r$ on the BLEU-4 score for the Book and Movie KG-Exp datasets. We average all top 10 runs for a more comprehensive comparison.}
\label{fig:lambda}
\end{figure}

As in~\cite{li-etal-2021-personalized}, we adapt the baseline codes to our setting and set the vocabulary size for NRT, ATT2Seq, and PETER to 20,000 by keeping the most frequent words. For PETER and PEPLER, we set the number of context words to 128. For all approaches, including KnowRec, we set the length of explanation to 128, as the mean length is about 94 for both datasets. %We keep all other baseline settings as their default. 
For KnowRec, we use an embedding size of 512, using a Byte-Pair Encoding (BPE) vocabulary~\cite{radford2019language} of size 50,256, with 2 encoding layers. Following KG generation work~\cite{ribeiro-etal-2021-investigating}, we split the tokens in the linearized graph with their corresponding label: \textit{[user], [graph], [head], [relation], and [tail]}. For both datasets, we set the batch size to 128 and max user and KG size to 64 and 192, respectively. We set the max node and edge length to 60. We experiment with $\lambda_r$ and $\lambda_e$ and find that 0.01 and 1 give us the best BLEU performance without affecting the recommendation prediction scores as in~\cite{li2022personalized}. See Figure~\ref{fig:lambda} for an analysis of Movie KG-Exp (Few-shot). The model's parameters were trained for 20 epochs and optimized via Adam~\cite{kingma2015adam} with a learning rate of 1e-3 and $Adam \ \epsilon$ of 1e-08, and the gradients were clipped at 1.0. All other attention-related hyper-parameters were the same as used in previous work~\cite{lewis-etal-2020-bart}. We decoded the text via beam search~\cite{hokamp-liu-2017-lexically} with a beam size of 5. Experiments were performed on NVIDIA RTX 3090 GPUs. We evaluate the model based on the validation set's total loss instead of BLEU score due to computational limitations, saving the top 10 models for testing, because the model with the least loss does not necessarily result in the best NLG metrics. 

Because of computation limitations, for evaluation purposes, we randomly sample and evaluate on 1\% of the test set, containing 4,491 and 1,456 samples for the Book and Movie datasets respectively. Note, that the size of the test set
%\jun{What do you mean by the pronoun \enquote{this}? Does it mean the size of test data?}\ac{Please see change. Thank you.}
is comparative to other text generative tasks such as KG-to-text~\cite{gardent-etal-2017-webnlg} and summarization~\cite{yu2022survey}. 

\subsection{Entity Coverage}
\label{sec:entity-coverage}
We define entity coverage (EC) as the percentage of unique entities, originating in an item KG, which appears in the recommendation explanation. More formally, for each head and tail entity \textit{e} in an item KG's set of entities \textit{E}, we calculate the token overlap in the explanation output for those entities. The EC score ranges in $[0, 1]$, where we report the percentage value in our results. The Book KG-Exp and Movie KG-Exp had an EC score of 71.45\% and 71.32\%, indicating that a descriptive, objective explanation should have a high EC score. The formula for EC is defined as:
\begin{align*}
\frac{\#KG\,entities\,found\,in\,output}{\#KG\,entities}
\end{align*}
or is the recall of the entities in a KG.

\section{Generated Examples}
\label{sec:examples}
% Ground Truth     |  Generated Sentences
% User-Item Graph  |
Table~\ref{tab:examples} presents some  examples generated by KnowRec from the Book and Movie KG-EXP datasets. As discussed in \cref{sec:analysis}, we find the examples to be fluent and grammatical, while incorporating both item features and implicit user information based on a user's purchase history. The generated examples closely match the ground truth, while integrating some language derived from the user. Note, that our aim here is to illustrate examples that showcase the implicit user preferences, instead of showing those generated outputs which most closely match the ground truth descriptions. As with other state-of-the-art NLG models, KnowRec does have a tendency to hallucinate by adding extra information that may not be necessarily accurate. As can be by the NLG metrics in \cref{tab:fullresults}, KnowRec relieves the hallucination problem by incorporating the user-item KG information. Such limitations may be additionally improved by leveraging more dense background KGs to generate from, while also incorporating user purchase history item features.
% \usepackage{booktabs}
% \usepackage{graphicx}
\begin{table*}[ht]
\centering
\small
% \resizebox{\textwidth}{!}{%
\begin{tabular}{@{}p{6.0cm}p{4.5cm}p{4.5cm}@{}}
\toprule
%\textbf{Knowledge Graph} &
%  \textbf{Generated} &
%  \textbf{Ground Truth Text} \\
%Knowledge Graph &
Item Graph Representation &
Generated Explanation &
Ground Truth Explanation \\
\midrule
 %\raisebox{-\height}{\includegraphics[width=\kgwidth]{figures/movie_1.pdf}} &
 \adjustbox{valign=t}{\includegraphics[width=3.9cm]{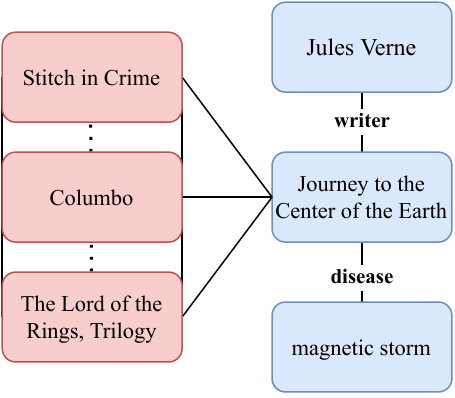}} &
  a scientist (\textbf{jules verne}) \underline{investigates} a \textbf{magnetic storm} that sends a  \underline{mysterious beam of light} from earth \textbf{to the center of earth}. &
  jules verne's professor lindenbrook leads a trip through monsters, mushrooms and a magnetic storm. \\ \midrule
 %\raisebox{-\height}{\includegraphics[width=\kgwidth]{figures/movie_2.pdf}} &
 \adjustbox{valign=t}{\includegraphics[width=3.9cm]{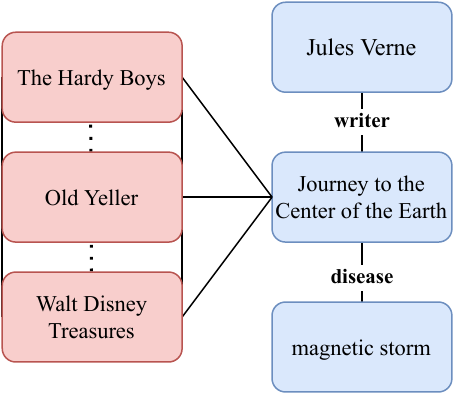}} &
  a group of scientists, inspired by \textbf{jules verne's} \underline{classic novel}, take a trip to the \textbf{magnetic storm} at \textbf{the center of the earth}. &
  jules verne's professor lindenbrook leads a trip through monsters, mushrooms and a magnetic storm. \\ \midrule
 %\raisebox{-\height}{\includegraphics[width=5cm]{figures/book_1.pdf}} &
 \adjustbox{valign=t}{\includegraphics[width=6.0cm]{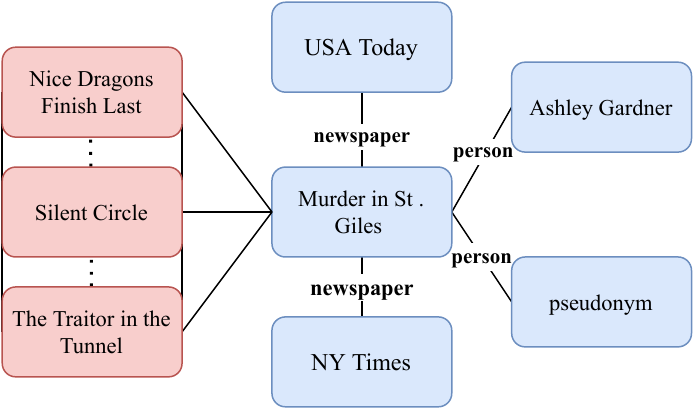}} &
  \textbf{ashley gardner} is a \textbf{ny times} and \textbf{usa today} bestselling author. under the \textbf{pseudonym} jennifer ashley, she has collectively written more than 70  \underline{mystery} and \underline{historical novels}. &
  usa today bestselling author ashley gardner is pseudonym for ny times bestselling author jennifer ashley. \\ \midrule
 %\raisebox{-\height}{\includegraphics[width=5cm]{figures/book_2.pdf}} &
 \adjustbox{valign=t}{\includegraphics[width=6.0cm]{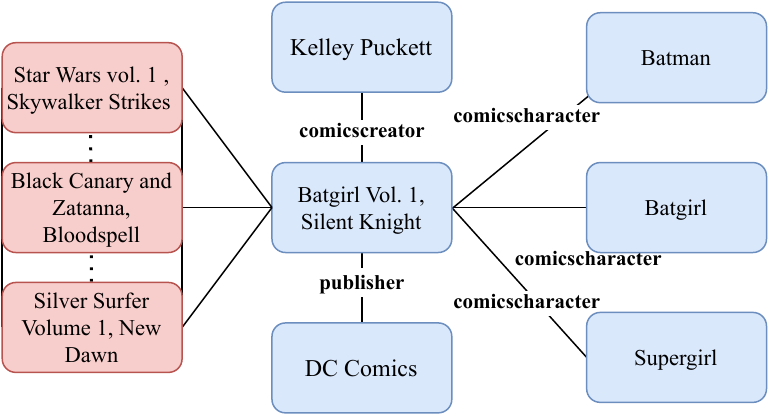}} &
  \textbf{kelley puckett} is an american comic book writer best known for his work on \textbf{batman} for \textbf{dc comics}.  he is the author of \underline{numerous books} for \underline{young readers}, including \textbf{supergirl}, the ultimate guide to character development and \textbf{batgirl}, a guide to writing for \underline{comics}, both published by image. &
  kelley puckett has been writing comics for far too long, by general consensus. he has worked on such series as batman adventures, batgirl and kinetic and supergirl for dc comics. \\ \midrule
 %\raisebox{-\height}{\includegraphics[width=5cm]{figures/movie_3.pdf}} &
 \adjustbox{valign=t}{\includegraphics[width=6.0cm]{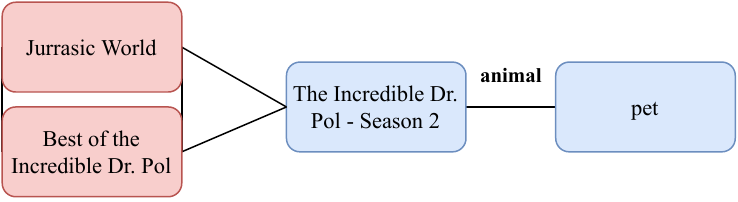}} &
  your \underline{favorite} \textbf{dr. pol} \underline{vet} and his \textbf{pet} dog \underline{return for a second season} of this hilarious and heartwarming animated adventure. &
  from sick goats to sick pet pigs, dr. pol and his colleagues have their hands full with a variety of cases and several animal emergencies. \\ \midrule
 %\raisebox{-\height}{\includegraphics[width=4.5cm]{figures/book_3.pdf}} &
 \adjustbox{valign=t}{\includegraphics[width=3.9cm]{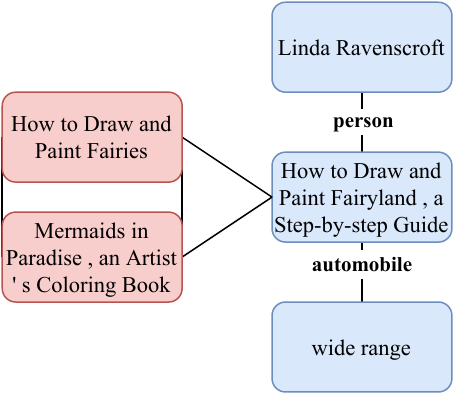}} &
  \textbf{linda ravenscroft} is an award-winning \underline{children's book} author and \underline{illustrator} who has illustrated a \textbf{wide range} of books and magazines, including the best-selling \underline{how to draw and paint series}. &
  linda ravenscroft has produced a wide range of images in fairyland motifs, including fine art prints, exclusive giftware, and fantasy art books. \\ \bottomrule
\end{tabular}%
% }
\caption{Examples generated by KnowRec on the Book/Movie KG-Exp datasets.
%We color code user nodes in red and item nodes in blue.
In the first column, we follow the format of user-item KG representation in \cref{fig:model}, where red nodes represent a user's  purchase history and blue nodes represent an item KG. For clarity and brevity, we only show the relevant parts of the item graphs.
In the second column,
the bold words
% in the text
are the item features directly coming from the item KG representation, whereas the underlined words are the features implicitly captured by KnowRec, based on the user's purchase history.}
\label{tab:examples}
\end{table*}

\end{document}